\documentclass[10pt,journal]{IEEEtran}

\ifCLASSOPTIONcompsoc
  \usepackage[nocompress]{cite}
\else
  \usepackage{cite}
\fi

\usepackage{amssymb}
\usepackage[cmex10]{amsmath}
\usepackage{mathabx}

\usepackage{xcolor,soul,framed} 

\usepackage[pdftex]{graphicx}
\graphicspath{{Figures/}}
\DeclareGraphicsExtensions{.eps,.jpg,.png,.pdf}

\usepackage{mdwmath}
\usepackage{mdwtab}
\usepackage{eqparbox}
\usepackage{booktabs}
\usepackage{url}

\begin{document}
  \title{Growing and Evolving 3D Prints}
  \author{Jon McCormack and Camilo Cruz Gambardella
  \thanks{Manuscript received November 12, 2020.}%
  \thanks{This research was supported by an Australian Research Council Grant FT170100033.}
  \thanks{Jon McCormack is Director of SensiLab, Monash University, Australia (e-mail: Jon.McCormack@monash.edu).}
  \thanks{Camilo Cruz Gambardella is a research fellow in SensiLab, Monash University, Australia (e-mail: camilo.cruz@monash.edu).}}%

\markboth{IEEE TRANSACTIONS ON EVOLUTIONARY COMPUTATION, AUTHOR PREPRINT 2021
}{McCormack \MakeLowercase{\textit{et al.}}: Growing and Evolving 3D Prints}


\IEEEtitleabstractindextext{%
\begin{abstract}
Design -- especially of physical objects -- can be understood as creative acts solving practical problems.
In this paper we describe a biologically-inspired developmental model as the basis of a generative form-finding system. 
Using local interactions between cells in a two-dimensional environment, then capturing the state of the system at every time step, complex three-dimensional (3D) forms can be generated by the system.
Unlike previous systems, our method is capable of directly producing 3D printable objects, eliminating intermediate transformations and manual manipulation often necessary to ensure the 3D form is printable. 
We devise fitness measures for optimising 3D printability and aesthetic complexity and use a Covariance Matrix Adaptation Evolutionary Strategies algorithm (CMA-ES) to find 3D forms that are both aesthetically interesting and physically printable using fused deposition modelling printing techniques. 
We investigate the system’s capabilities by evolving and 3D printing objects at different levels of structural consistency, and assess the quality of the fitness measures presented to explore the design space of our generative system. We find that by evolving first for aesthetic complexity, then evolving for structural consistency until the form is `just printable', gives the best results. 
\end{abstract}

}

\maketitle

\IEEEdisplaynontitleabstractindextext

\section{Introduction}
Architects and Designers sit at the intersection of engineering and art, because designed artefacts provide both practical solutions to everyday problems and have an aesthetic impact that shapes the environment in which they operate \cite{hillier1989social, lawson1990designers}.
Often these two aspects of design (addressing practical problems and making a cultural impact) are in conflict, typically due to combining unconnected approaches, such as formal optimisation and purely intuitive craft. This leads to addressing the practical and aesthetic considerations separately, often resulting in poor or disjointed design solutions.

One family of methods that have been in use in the arts since Aristotle are generative systems: techniques that use the decomposition, analysis and reorganisation of existing artefacts to give way to new ones \cite{Mitchell1977}. Amongst them, self-organising simulation methods, such as cellular automata \cite{Adamatzky2016}, developmental models \cite{McCormack2005c}, agent-based models \cite{GreenfieldM09,Jones2015} and evolutionary methods \cite{Bentley1999,BentleyCorne2002} have been adopted as a means to explore creative, aesthetically driven spaces in contemporary architecture, art and design.

These systems have proven particularly useful in developing complex, aesthetically novel forms \cite{lynn1999animate, Porter2010, dino2012creative}, but with a priority on the aesthetic aspects of the object, leaving the practical aspects to be solved through traditional and largely human-driven design optimisations.
Evolutionary optimisation methods are often used to solve practical design problems, but usually come into the process after the aesthetic design decisions have been made. Typically they address quantifiable problems, such as the distribution of structural elements \cite{munk2015topology} or optimisation of energy consumption \cite{waibel2019building}.

Hence, an exploration into integrated systems capable of reconciling aesthetic and practical design goals can contribute significantly to advancing computational design practice.

We present the design of a generative system that incorporates methods to simultaneously evaluate both pragmatic and creative value, seeking to balance the tension between aesthetics (visual complexity) and practical goals (fabrication). Our aim is to bring the digital and material qualities of design into closer harmony, allowing the designer to explore the aesthetic possibilities of complex generative systems and the material possibilities of design through digital fabrication. 

Our system has its origins in the layering process of additive manufacturing: synthesising physical objects directly through the use of 3D printing technologies \cite{Redwood2018}. Since toolpath manipulation is the basis of form generation in our system, we bypass the traditional problem of transforming 3D digital mesh models into 3D printer toolpath operations using techniques such as slicing and scaffolding \cite{Steuben2016} -- often a challenging process that can result in incomplete or failed prints \cite{Redwood2018} and which omit certain creative possibilities for printing, such as control over a single layer of filament deposition. The main limitation of this approach relates to printability; something we address through a formalised fitness measure.
We then deploy evolutionary methods that combine practical and aesthetic fitness measures to allow an efficient exploration of the design space by automatically rejecting designs that are infeasible to fabricate or aesthetically poor.

We adopt a ``generate and explore'' methodology, similar to the one designers intuitively use. Motivated by the inherent design complexity of biological systems, and the need to directly control the 3D printer toolpath, our system employs a combination of physics and biology simulation to generate complex 3D forms, then a CMA-ES algorithm for exploration of the feasible design space. We introduce metrics to evaluate aesthetic complexity and structural consistency, allowing the designer to explore the tension between design possibility and fabrication integrity.

The remainder of this paper is structured as follows: first, an overview of work related to generative and optimisation systems in design is provided (Section \ref{s:relatedWork}). Next, the two main components of the proposed system -- the generative developmental system (\ref{s:form-generation-model}) and the form-finding exploratory system (\ref{ss:es-form-finding}) -- are outlined. Section \ref{s:experiments} describes the experimental setup in detail, and quantitative results are presented, alongside the 3D prints of the grown physical objects evolved by the system. Section \ref{s:analysis} reflects on the results produced by our method, along with suggestions for future development before concluding (Section \ref{s:conclusion}).

\section{Related Work}
\label{s:relatedWork}

\subsection{Generative Systems in Art and Design}
\label{ss:generativeArt}

Generative art has a history that predates computers by thousands of years \cite{mccormack2014ten} and generative techniques have been well studied as a framework for art and design \cite{Whitelaw2005,McCormack2004b,BodenEdmonds2009,reas2010,Dorin2012,Bohnacker2012}. We limit our examination to computational examples of 3D form generation using biologically-inspired techniques that directly relate to the research presented here.

The irruption of computers into design practice and research has lead to a multitude of generative approaches to design, ranging from form exploration to structural optimisation. Early adopters looked at natural processes \cite{Stevens1974}, patterns of organisation and development \cite{Alexander1964, alexander1977pattern} or formal design grammars based on shape \cite{Stiny1975}. More recent forays borrow from advances in complex systems, artificial life and agent-based models \cite{McCormack2004b}.

A notable example of this kind of approach is \textit{Accretor}, a project in which Dutch artists Driessens and Verstappen used a semi-totalistic 3D cellular automaton to generate a range of diverse form configurations, from regular and `boring', to highly disordered \cite{whitelaw2015accretor}. Computationally generated forms were human-selected for 3D printing, searching for what the authors refer to as \textit{in between objects}. The criteria used for categorisation of the outcomes was not disclosed, but was likely based on the artists' aesthetic preferences.

Similarly, digital artist Andy Lomas' work \textit{Cellular Forms} explores the morphogenetic capabilities of a generative system based on cellular reproduction \cite{lomas2013, lomas2014cellular}. The process begins with a single spherical cell that progressively subdivides based on environmental conditions. Lomas used principles analogous to photosynthesis to generate sophisticated and diverse biomorphic 3D forms, a number of which were 3D printed. Initially the materialisation was achieved ``manually'' through standard 3D printing slicing and scaffolding, but in more recent work Lomas has used layer overhang metrics to automate determination of successful fabrication \cite{Lomas2019}.

\subsection{Evolutionary Methods}
\label{ss:evolutionaryMethods}

Many generative artists and designers (see, e.g.\ %
\cite{Sims1991,McCormack1992,Todd1992, Bentley1999, BentleyCorne2002}) employ evolutionary methods as a search or optimisation strategy, often a variant of the \emph{Interactive Genetic Algorithm} (IGA) \cite{Takagi2001}. The IGA is popular for creative applications because it substitutes formalised fitness measures with human judgement. It circumvents the difficulty of developing generalised fitness measures for ``subjective'' criteria, such as personal aesthetic judgement or taste, allowing users to move through a design space, hopefully guiding the evolutionary search towards specific aesthetic results. However, problems with the IGA are well known: human evaluation creates a bottleneck; subjective comparison is only possible for a small number of individuals (i.e.\ $< 20$); human users become fatigued after only a few generations; evolving to specific targets is often difficult or impossible. There have been many attempts to overcome these problems, for example distributed evolution with multiple users \cite{Secretan2011}, but distributed techniques are obliviously incompatible with individual designers or personal aesthetic preferences. In the basic IGA the role of the human creative is the equivalent of a ``pigeon breeder'', where the only valid action is selecting from a small set of individuals. This may be why the method has been more suited to use by non-experts than design professionals\footnote{Some commercial design and music software products and have integrated IGAs or similar evolutionary methods into design workflows, but in general the technique has not been adopted professionally.} \cite{Takagi2001,bownMcCormack2010}.

Another way to address problems with the IGA is to formalise or automate the aesthetic evaluation, freeing the human user from endless comparative evaluations. Research communities from both computational aesthetics (CA) and psychology have proposed numerous theories and measures of aesthetics \cite{Johnson2019}. However, a computable, universal aesthetic measure remains an unsolved open problem \cite{McCormack2005a}. One of the main barriers is the psychological nature of aesthetic judgement and experience. Leder et al.'s model of aesthetic appreciation and judgement describes information-processing relationships between various components that integrate into an aesthetic experience and lead to an aesthetic judgement and aesthetic emotion \cite{Leder2004,Leder2014}. The model includes perceptual aesthetic properties, such as symmetry, complexity, contrast, and grouping, but also social, cognitive, contextual and emotion components that all contribute to forming an aesthetic judgement. A key element of Leader's revised model \cite{Leder2014}, is that it recognises the influence of a person's affective state on many evaluative components and that aesthetic judgement and aesthetic emotion co-direct each other. Hence, a comprehensive computational aesthetic measure must consider the interaction between cognition and affect, along with prior knowledge and viewing context of the human observer, making generalisation extremely difficult.

Neural networks have long been proposed as a way to automate aesthetic judgement in creative evolutionary systems \cite{Baluja1994}. Recent approaches rely on advances in deep learning to learn individual artistic or stylistic preferences \cite{McCormackLomas2020,McCormackLomas2020b}. While these methods have had some success, they require a large training corpus that has user-defined rankings or categorisations, which is typically a manual, time-intensive process.

An important difference in the work introduced here is that we formalise both aesthetic and practical measures to assist in design generation, but we do not remove the human designer completely from the design process. We automate fitness evaluation but allow the designer to dynamically change the weighting between aesthetics and practicality to explore the design space.

\subsection{Generative 3D printing}

Researchers have explored applications of generative and evolutionary systems to solve practical problems related to 3D printing. In \cite{tutum2018functional}, the authors use a combination of generative and evolutionary techniques to produce mechanically functional objects that can be manufactured using a consumer-grade 3D printer. Similarly, \cite{yu2017evolutionary} use decomposition of digital 3D models and evolutionary approaches (CMA-ES and MOEA) to optimise the manufacturing process by reducing the use of waste material. Techniques developed in \cite{zhao2016connected} introduced the use of space-filling curves for 3D printing, as an efficient alternative to infilling. The method defines a single line toolpath, minimising material usage. These techniques are aimed at printing optimisation for pre-existing 3D designs, the difference in our approach is that we integrate printing optimisation and aesthetic design as a single process, directly outputting toolpath instructions (as opposed to geometry) to the 3D printer.

As discussed in the previous section, recent work by Lomas examined automated methods to eliminate overhangs in morphogenetic 3D printed forms \cite{Lomas2019}. He developed a set of constraints that were built into the form generation system to prevent growth that would likely result in an unprintable form.

\begin{figure}
    \centering
    \includegraphics[width=0.95\columnwidth]{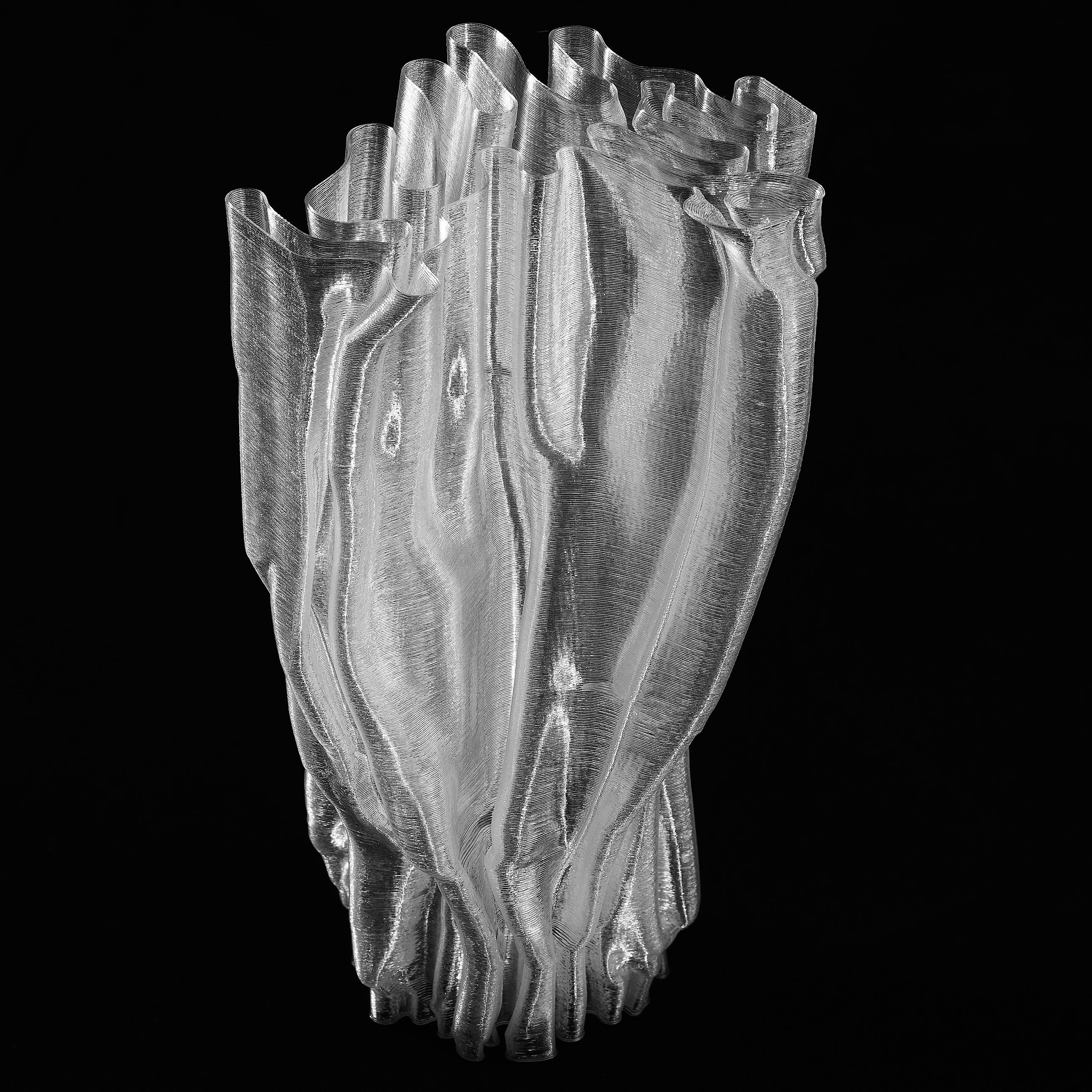}
    \caption{A 3D printed form grown using the system described in this paper}
    \label{fig:heroForm}
\end{figure}

Our method is concerned with maintaining toolpath profile generation within a usable range via a computable fitness measure (Figure \ref{fig:heroForm} shows an example). We build a 3D form through accreted vertical layering, where the major concern is the physical change between consecutive layers: without sufficient material in any preceding layer to support the next, the print will likely fail. This requirement often sits in tension with our aesthetic requirements that are based on variations in complexity of the final form. We allow the designer an exploratory trade-off between design aesthetics and risk of the form being unprintable.

So while the work we present in the following sections shares principles with the projects mentioned in this section, it advances these ideas by combining a generative system with an evolutionary one to explore a design space where practical and aesthetic characteristics are considered as equally valuable, and viable designs emerge through a process where practical and aesthetic possibilities are co-developed. 

\section{Form Generation Model}
\label{s:form-generation-model}

Our model consists of two main components: (i) a generative developmental system for form generation (described in this section), and (ii) an exploratory evolutionary system for form finding (described in Section \ref{ss:es-form-finding}). The source code implementation of our model is available on-line \cite{Cruz2021}.
The generative developmental system uses a physically-based, biologically-inspired model that simulates the development of an organic form via morphogenesis. The model has a series of encoded parameters (the \emph{genotype}, see Section \ref{ss:genetic-parameters}) that affect the form's developmental process and hence, the resultant form.\footnote{Other factors, such as the simulated developmental environment also play an important role, we discuss this further in Sections \ref{ss:dgs} and \ref{ss:food}.} The exploratory system allows efficient searching through the design space of this developmental system. Through the use of a weighted, multi-objective fitness function, the exploratory system aims to search for forms that are both aesthetically interesting and directly printable using FDM 3D printing.

Our generative developmental system is based on Differential Growth (DG) in two dimensions \cite{barlow1989differential}. The motivations for using this system include its ability to coherently generate complex forms through development and the 2D constraint allows us to conveniently layer each developmental cycle directly as a 3D print layer. We use the metaphor of cells connected in a sequence to form an ``organism''. Cells seek to acquire energy from the environment in order to grow and reproduce. Organisms may also divide into two or more separate entities as part of their development. We capture this developmental process over time and translate it directly to 3D printer toolpath instructions, building a finished print a series of stacked layers (Figure \ref{fig:heroForm}), without recourse to intermediate representations or the need for extraneous scaffolding or support structures often needed in 3D printing.

Biology serves as the inspiration in the form generation process, because nature is a rich source of design ideas, capable of generative complexity and functional diversity \cite{ball1999self}. However, it is important to emphasise that the model is not meant to inform biology or be physically or biologically faithful -- biology is the inspiration but design is the goal.

\subsection{Developmental System} 
\label{ss:dgs}

Development in our system occurs in discrete, fixed timesteps ($t_{0}, t_{1}, \ldots t_{n}$, with $t_{i+1}-t_{i} = \Delta t$) in a physically simulated two-dimensional, viscous medium of uniform density (the \emph{environment}). Capturing the system state at each timestep is an important part of the form generation process (detailed in Section \ref{ss:simulation-into-form}). 

We define an \emph{organism} ($\mathcal{O}$) as a series of connected \emph{cells} $\mathcal{C}_1 \ldots \mathcal{C}_n$. A cell, $\mathcal{C}_i = (v,m,\varepsilon) \in \mathbb{R}^2 \times \mathbb{R}_{+} \times \mathbb{R}_{+}$, where $v \in \mathbb{R}^2$ is the cell's \emph{vertex} (location), $m \in \mathbb{R}_{+}$ the cell's mass and $\varepsilon \in \mathbb{R}_{+}$ the cell's energy.
Each cell is connected in sequence by \emph{edges}, $e_1 \ldots e_n$, where $e_i = \langle v_i,v_{i+1} \rangle$ for $1 \le i < n$ and $e_n = \langle v_n, v_1 \rangle$, ensuring closed loops. Edges use a simulated tension and compression spring to help maintain separation of their connected cells, and also to prevent them from moving too far apart. Physical attributes of both the cells and edges determine the growth and change in shape over time.\footnote{Our system was inspired by Anders Hoff's differential line algorithm: \url{https://inconvergent.net/generative/differential-line}}
The full set of organisms generated over time $<t_0 \ldots t_n>$ is called a \emph{colony}, $\mathcal{Y}_n$. The colony is the unit of selection in the evolutionary component of our system.

\subsection{Genetic Parameters}
\label{ss:genetic-parameters}

A colony shares a genome, $G \in \mathbb{R}^5$ with the following alleles:
\begin{itemize}
        \item \textbf{Metabolic rate} ($\eta$)  determines a cell's capacity to transform resources into energy, i.e.~how efficiently  a cell converts acquired nutrients into usable energy (Section \ref{ss:food}). Cells that metabolise more efficiently also consume more energy, and vice-versa.  
        \item \textbf{Cell drag coefficient} ($\nu$) represents the surface drag of a cell per unit area. Low drag allows the cell to move more quickly but makes stable configurations more difficult to achieve.
        \item \textbf{Energy capacity} ($\varepsilon_{max}$) defines a cell's capacity to store energy. Greater capacity increases cell mass, increasing the amount of energy required for movement and metabolism.
        \item \textbf{Edge spring coefficient} ($k$) determines edge stiffness of the springs between cells (see Section \ref{ss:dgs}). High ($\ge 1$) coefficients give stiff connections, resulting in organisms that appear rigid. Lower coefficients allow cells to move more freely, giving way to organisms that move more fluidly. The disadvantage of low spring coefficient is that connections tend to be longer, which makes the transmission of energy between cells less effective, as more energy is lost in transmission. Very low coefficients make it difficult for cells to move effectively to locate nutrients, as they tend to overshoot the source. 
        \item An \textbf{Energy ratio} ($\rho$) controls the proportionate size of new organisms after the splitting process occurs.
\end{itemize}

Together these genetic parameters determine the final 3D form generated by the development process. We next discuss the physical aspects of the model.

\subsection{Physical Simulation}
\label{ss:physical-simulation}

\begin{figure}
    \centering
    \includegraphics[width=0.95\columnwidth]{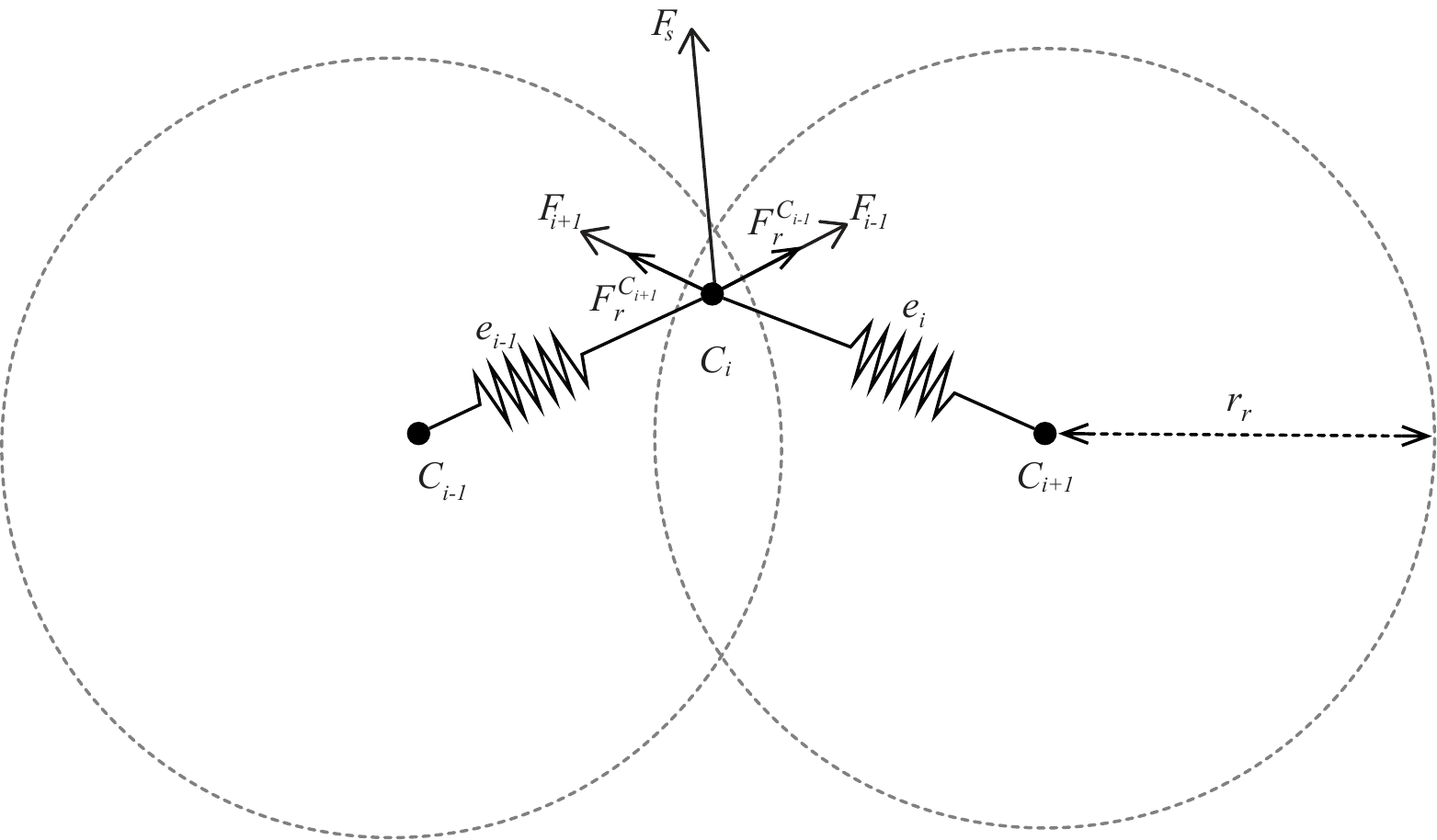}
    \caption{Cells connected by edges with springs: forces shown acting on a single cell. $F_{i+1}$ represents the sum of forces excerpted on cell $C_i$ by its neighbouring cell $C_{i+1}$ and the spring that connects them $e_i$, $F_{i-1}$ represents the sum of the force excerpted on $C_i$ by neighbouring cell $C_{i-1}$ and their connecting spring $e_{i-1}$, and $F_s$ represents the sum of repulsion forces by all the other cells in the system closer than $r_r$ to $C_i$}
    \label{fig:forces}
\end{figure}

To help maintain cell and organism separation, cells exert a repulsive ($F_r$) force on every other cell within a specified radius from the cell's centre, $r_r$ (Figure \ref{fig:forces}). The repulsion force between two cells $\mathcal{C}_i$ and $\mathcal{C}_j$ is applied according to:

\begin{equation}
     F_r = \begin{cases}
        k_r \left( -1/d_{i,j}^2 + 1/r_{r}^{2} \right) & \text{if } d_{i,j} \le r_r \\
        0 & \text{otherwise}
        \end{cases}
\end{equation}
where $d_{i,j}$ is the distance between the centres of $\mathcal{C}_i$ and $\mathcal{C}_j$ and $k_r$ a constant used to control the relative amount of repulsive force. Additionally, spring forces also act on connected cells based on Hooke's law: $F_s = -k_s \delta$, where $k_s$ is the spring coefficient and $\delta$ is the compression (-ive) or expansion (+ive) distance of the spring over its \emph{rest length}, $l^{e_i}_{t}$ (Figure \ref{fig:forces}). Rest length is a dynamic property of an edge, $e_i$, dependent on cell energy and explained in the next section. Cells can also instigate motion towards a positive nutrient gradient using internal forces, this is detailed in Section \ref{ss:food}. A cell's mass is proportional to its energy (Section \ref{ss:energy}). At each timestep, the total forces on each cell are summed and the cell moves as a result of the cumulative forces acting on it.

\subsection{Energy Model}
\label{ss:energy}

Each cell, $\mathcal{C}_i$, has an \emph{energy}, $\varepsilon$, representing the metabolised energy currently stored in that cell. Cells are created with a default energy, $\varepsilon_{init}$, and may accumulate, diffuse and loose energy as the organism develops. The maximum energy a cell can accumulate ($\varepsilon_{max}$) is genetically determined (Section \ref{ss:genetic-parameters}). 
Let $E : \texttt{cell}, t \to \mathbb{R}$ be a function that returns the energy level of a given cell $\mathcal{C}_i$ at time $t$, and $E :  \texttt{Organism}, t \to \mathbb{R} = \sum_{i=1}^{n} E(\mathcal{C}_i^\mathcal{O},t)$ a function that returns the total energy for all cells in $\mathcal{O}$ at time $t$. The overall change in energy for a cell at each timestep is given by:
\begin{equation}
\label{eq:energy}
    E(\mathcal{C}_i,t+1) = E(\mathcal{C}_i, t) - \mathcal{E} + \mathcal{D} + \mathcal{N} 
\end{equation}
Where $\mathcal{E}$ represents internal energy loss, $\mathcal{D}$ energy diffused and $\mathcal{N}$ energy metabolised.
The energy loss per timestep $\mathcal{E}$ is calculated as:
\begin{equation}
    \mathcal{E} = c_1 + \eta \cdot (\varepsilon_{max} + c_2 \frac{F_N}{\nu^2}),
\end{equation}
\noindent
where $c_1$ is a `cost of living' constant, $\eta$ is the cell's metabolic rate, $\varepsilon_{max}$ is its energy capacity (Section \ref{ss:genetic-parameters}), $c_2$ a constant and $F_N$ the internal movement force (explained further in Section \ref{ss:food}).

A diffusion process ($\mathcal{D}$) occurs at each timestep, whereby energy from cells with higher energy is diffused to those neighbours with lower energy. For example, if $E(\mathcal{C}_i) >  E(\mathcal{C}_{i+1})$ then:
\begin{equation}
  \begin{gathered}
    \begin{aligned}
  \mathcal{D}(\mathcal{C}_{i+1},t) &= \chi\cdot  E(\mathcal{C}_{i},t) - (\omega\cdot d) \  \text{and} \\ \mathcal{D}(\mathcal{C}_{i},t) &= - \chi \cdot E(\mathcal{C}_{i},t),
    \end{aligned}
    \end{gathered}
\end{equation}
where $\chi$ is the diffusion rate, $\omega$ is the loss of energy per unit distance (both constants), and $d$ is the length of the edge connecting $\mathcal{C}_i$ and $\mathcal{C}_{i+1}$. This diffusion process `evens out' the energy distribution of the organism over time, helping cells that can't directly acquire energy to survive.
An organism grows and develops through the accumulation of energy acquired by cells from nutrients in its environment, $\mathcal{N}$ (detailed in Section \ref{ss:food}).

Energy determines how a cell changes, according to the following condition-action rules:
\begin{equation*}
 \begin{split}
    \texttt{if }  E(\mathcal{C}_i) = 0 & \texttt{ then } Die \\
    \texttt{if }  E(\mathcal{C}_i) \geq \varepsilon_{max} & \texttt{ then } Divide
    \end{split}
\end{equation*}

The $Die$ action occurs when a cell's energy level becomes 0. The cell and its leading edge are deleted from the organism and the previous cell's edge ($e_i$) connected to the successor cell. The rest length of connecting springs is determined proportional to the sum of the energies of the cells at each end of the edge: 
\begin{equation}
\label{eq:rest-length}
    l_{e_i} \propto \sqrt{ E(\mathcal{C}_{i}) + E(\mathcal{C}_{i+1})}
\end{equation} 

At least three cells are required to maintain a closed loop, so if less than three cells remain the organism is deleted.
The rest length of a spring connecting two cells changes according to the energy of the cells, as per (\ref{eq:rest-length}). So the more energy the longer the rest length, which has the effect of pushing high energy cells further apart.
The logic is based on a metaphor of metabolised growth. Growth increases the length of edges, effectively causing the organism to encompass a larger area, increasing its mass and surface area, which in turn impacts on the cell's dynamics.

\begin{figure}
    \centering
    \includegraphics[width=0.45\textwidth]{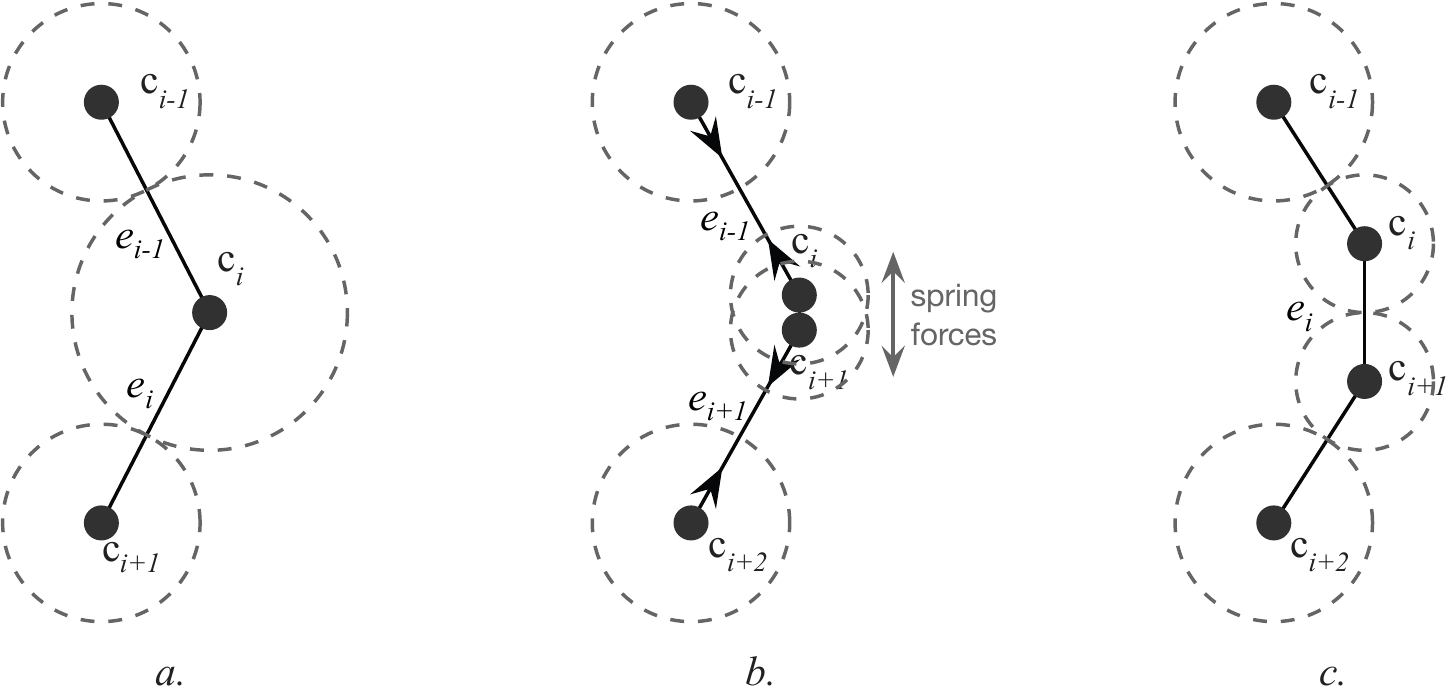}
    \caption{Cellular division: (a) Cell $c_i$ has reached $\varepsilon_{max}$ so will divide. (b) $c_i$ has divided into two new cells, $c_i$ and $c_{i+1}$. The rest lengths of the springs at each edge are set proportional to the cell's energy, causing the new cells to push apart until they reach equilibrium (c). The dotted circles represent the energy level of each cell.}
    \label{fig:cellular-division}
\end{figure}

When a cell's energy equals or exceeds $\varepsilon_{max}$ the $Divide$ action is initiated, which causes the cell to divide, spawning a new cell (Figure \ref{fig:cellular-division}).
This new cell receives half of the energy of its parent, and both cells share the same location at the start of the process. A new spring is created to connect the cells. The spring's rest length is determined by the energy of the two connecting cells (\ref{eq:rest-length}). 
The rest length causes energy minimisation in the system, pushing new cells apart until their connecting edges' rest lengths are achieved.

In biology, organisms acquire energy for growth through, e.g., photosynthesis or the digestion and metabolism of nutrients. We experimented with a number of different scenarios, finding the nutrient metabolism model the most expressive in terms of model form variation. This model in detailed in Section \ref{ss:food}.

\subsection{Organism Division}
\label{ss:organism-reproduction}
\begin{figure}
    \centering
    \includegraphics[width=0.4\textwidth]{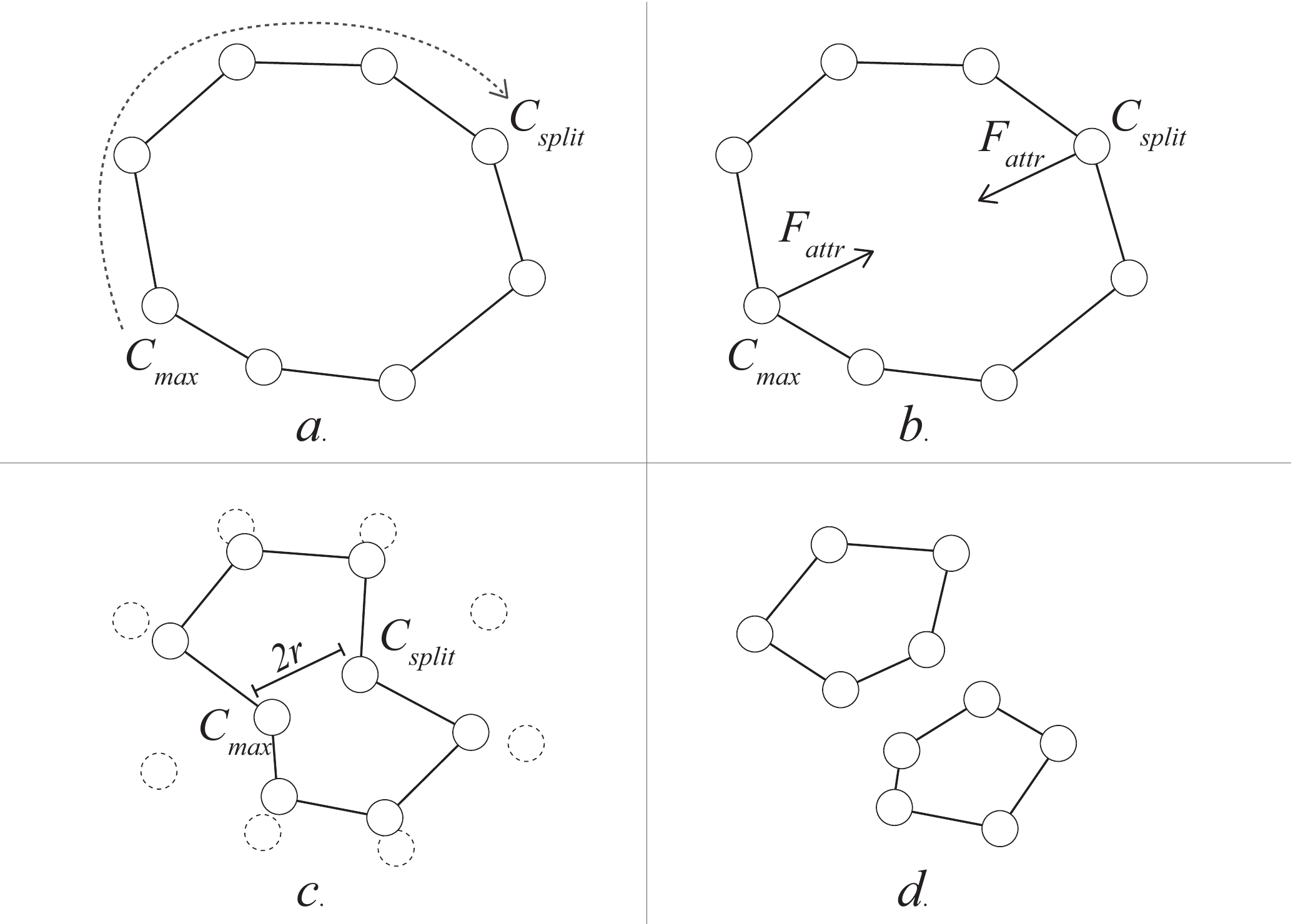}
    \caption{Stages of organism splitting. a) $\mathcal{C}_{max}$ and $\mathcal{C}_{split}$ are defined. b) Attraction forces between $\mathcal{C}_{max}$ and $\mathcal{C}_{split}$ are applied. c) Distance between $\mathcal{C}_{max}$ and $\mathcal{C}_{split}$ reaches splitting threshold ($2r_r$) d) Organism splits, producing two separate and independent organisms}
    \label{fig:org_split}
\end{figure}

At each timestep, the combined energy of all an organism's cells is calculated using $E(\mathcal{O},t)$.
If this total exceeds a threshold, $\varepsilon_o = 10 \cdot \varepsilon_{init}$, division occurs, with the original organism splitting into two. The division process proceeds in several steps (see Figure \ref{fig:org_split}): 
\begin{enumerate}
 \item the cell with the highest energy is found (i.e.~$\displaystyle \max_{1 \leq i \leq n} E(\mathcal{C}_i,t)$), let this cell be $\mathcal{C}_{max}$;
 \item beginning at $\mathcal{C}_{max}$ the organism is traversed along the leading (clockwise) edge, summing the energy of the next cell, $E(\mathcal{C}_{max + 1})$;
 \item the process continues until the ratio of summed energy to total energy ($E(\mathcal{O},t)$) equals $\rho$, let this cell be $\mathcal{C}_{split}$;
 \item An attractive force, $F_{attr}$, is applied between $\mathcal{C}_{max}$ and $\mathcal{C}_{split}$ until the distance between their centres is $\leq 2 r_r$;
 \item the cells $\mathcal{C}_{max}$ and $\mathcal{C}_{split}$ are duplicated and the leading edges of each are connected to the incoming and leading edges of their respective neighbour cells (see Figure \ref{fig:org_split}).
 \item the organism has split into two separate organisms and development of each now continues independently.
\end{enumerate}

Through these processes of cellular and organism growth, the system develops over time. Of course, growth requires energy, so we next outline the metabolic energy model.

\subsection{Metabolic Model}
\label{ss:food}
To provide energy to the growing cells we implemented a simple nutrient and metabolism model. We place a fixed number ($n_S$) of \emph{nutrition sources}, $S_1\ldots S_{n_s}$, with uniform random distribution within the 2D simulation environment.\footnote{Note that the overall size of the environment is determined by the 3D printer's build plate area, see Section \ref{sss:environment}.} Each nutrient source has a fixed production rate, $\alpha_{S_i}$, measured in nutritional units per timestep, $\Delta t$, and a \emph{production capacity} of $\beta_{S_i}$ nutritional units.

Nutrients are emitted from a source at rate $\alpha_{S_i}$ until the total nutrient production from the source $S_i$ reaches $\beta_{S_i}$. Once a source's production is exhausted, it is replaced by a new source, $S_j$ at a new random location in the environment. The constants $\alpha_{S_i}$ and $\beta_{S_i}$ are assigned from a normal distribution at the instantiating of $S_i$.

Emitted nutrients diffuse uniformly from the source location into the environment. Let the concentration of nutrients at a particular location in the environment be $N(x,y)$, then the diffusion of nutrients is given by:
\begin{equation}
    \dot{N} = \gamma^2\nabla^2 N - c_N = \gamma\left( \frac{\partial^2N}{\partial x^2}+\frac{\partial^2 N}{\partial y^2}\right) - c_N
\end{equation}

$\dot{N}$ is the time derivative of $N$, $\gamma$ is a global constant that controls the rate of diffusion and $\nabla^2 N$ is the Laplacian of $N$ and $c_N$ is a constant decay rate, representing nutrient loss to the environment. In practice we use a two dimensional array of discrete samples mapped over the environment and approximate $\dot{N}$ using finite difference methods.

Cells are able to perceive the presence of nutrients if they are within a fixed distance from the cell. When a cell senses nutrients it excerpts a force, $F_N$, in the direction of maximum positive gradient of nutrient. The magnitude of the force is proportional to $E(\mathcal{C})$ and exerting this force depletes a cell's energy at a rate $\propto F_N$ (see Section \ref{ss:energy}).

\subsection{Model Analysis}
\label{ss:modelAnalysis}
To analyse the contribution of each model component described in this section, we turned off components and generated 100 individuals with random seeding for each. Table \ref{tab:analysis} gives quantitative results of each component of the model. The table shows the normalised standard deviations relative to the full model, with $\sigma_P$ and $\sigma_C$ the standard deviation of printability and complexity fitness measures (Section \ref{ss:fitness-functions}), $\sigma_V$ volume and $\sigma_D$ volume variation. Values are shown with the energy model (Section \ref{ss:energy}), organism division (Section \ref{ss:organism-reproduction}), metabolism (Section \ref{ss:food}) and physical simulation (Section \ref{ss:physical-simulation}) ablated. Removing these model features generally results in significantly less variation in the diversity that the system is capable of generating.

\begin{table}[htbp]
    \centering
    \begin{tabular}{c|cccc}
   & $\sigma_P$ & $\sigma_C$ & $\sigma_V$ & $\sigma_D$ \\
   \midrule
   full model & 1.0 & 1.0 & 1.0 & 1.0 \\
   no energy & 0.637489	& 1.25948 &	0.724754 &	1.10708 \\
   no division & 0.47171 &	1.18837	& 0.523708 &	0.851265\\
   no meta & 0.0 & 0.144783 & 0.0193064 & 0.018196 \\
   no physics & 0.005308 &	0.384408 &	0.101377 &	0.281548\\
   \bottomrule
    \end{tabular}
    \caption{Normalised standard deviations for model components.}
    \label{tab:analysis}
\end{table}

\subsection{Simulation into Form}
\label{ss:simulation-into-form}

To turn our simulation into form we work within a set of physical constraints defined by 3D printing. The objective is to make this process as efficient as possible by directly controlling the deposition of physical material from the simulation without processing intermediate 3D geometric representations. To generate a 3D form that can directly control the 3D printer, we output the simulation geometry to the printer at each timestep, $t_i$. This is achieved by converting the simulation geometry directly into G-code,\footnote{G-code, also known as \emph{RS-274} is a widely used programming language for computer numerical control of computer aided manufacturing tools.} i.e.~there is no intermediate 3D representation as would normally be required when printing 3D objects. The G-code required to print in extrusion-based printers represents a series of movements between coordinates with instructions to control the volume of material extrusion when moving. As the simulation effectively consists of a series of connected lines, we use this information to directly drive the 3D print head in 2D, allowing the printer to deposit a layer of thin filament that represents organisms' shapes under simulation at each time step.

\begin{figure*}
    \centering
    \begin{tabular}{cccc}
    \includegraphics[width=0.22\textwidth]{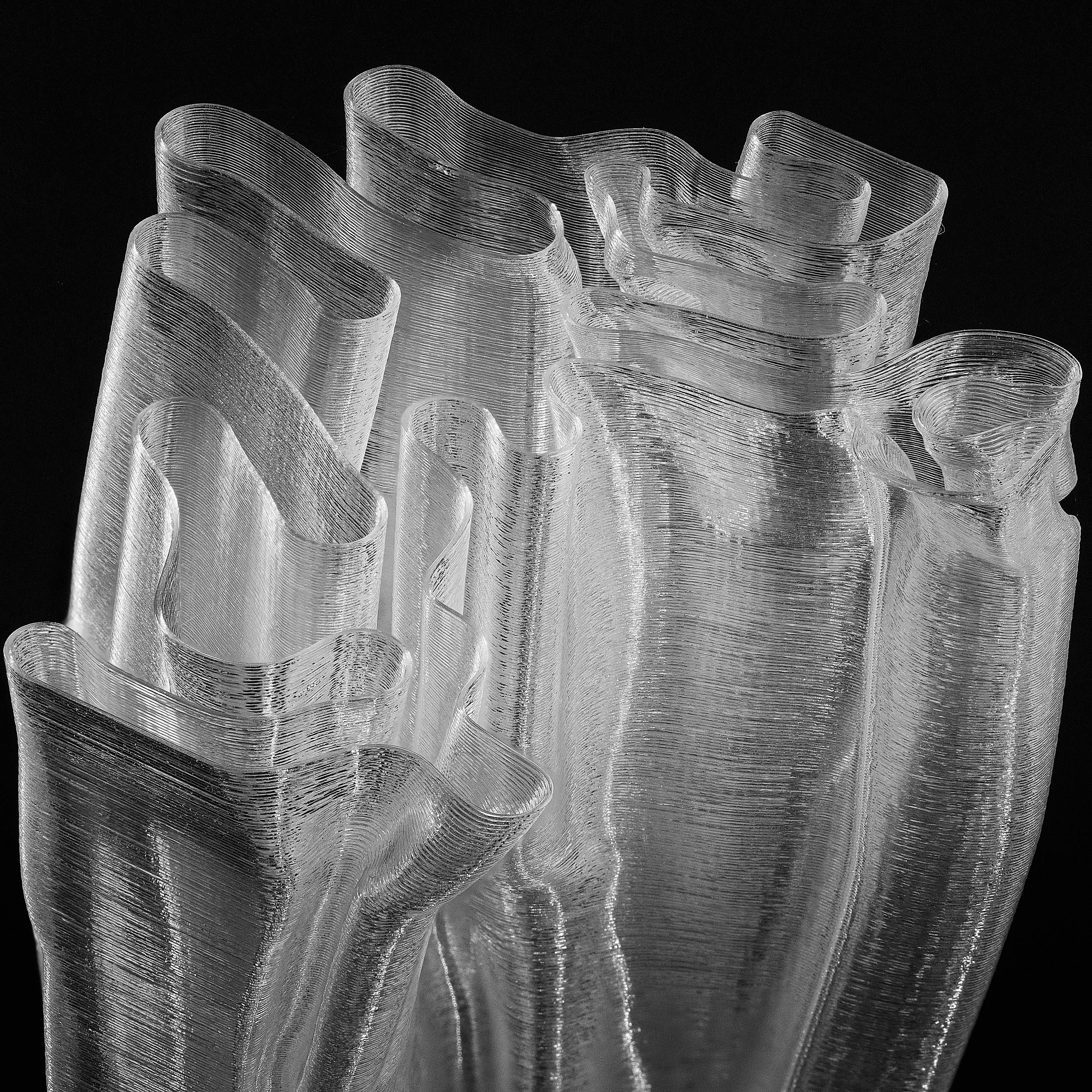} &
    \includegraphics[width=0.22\textwidth]{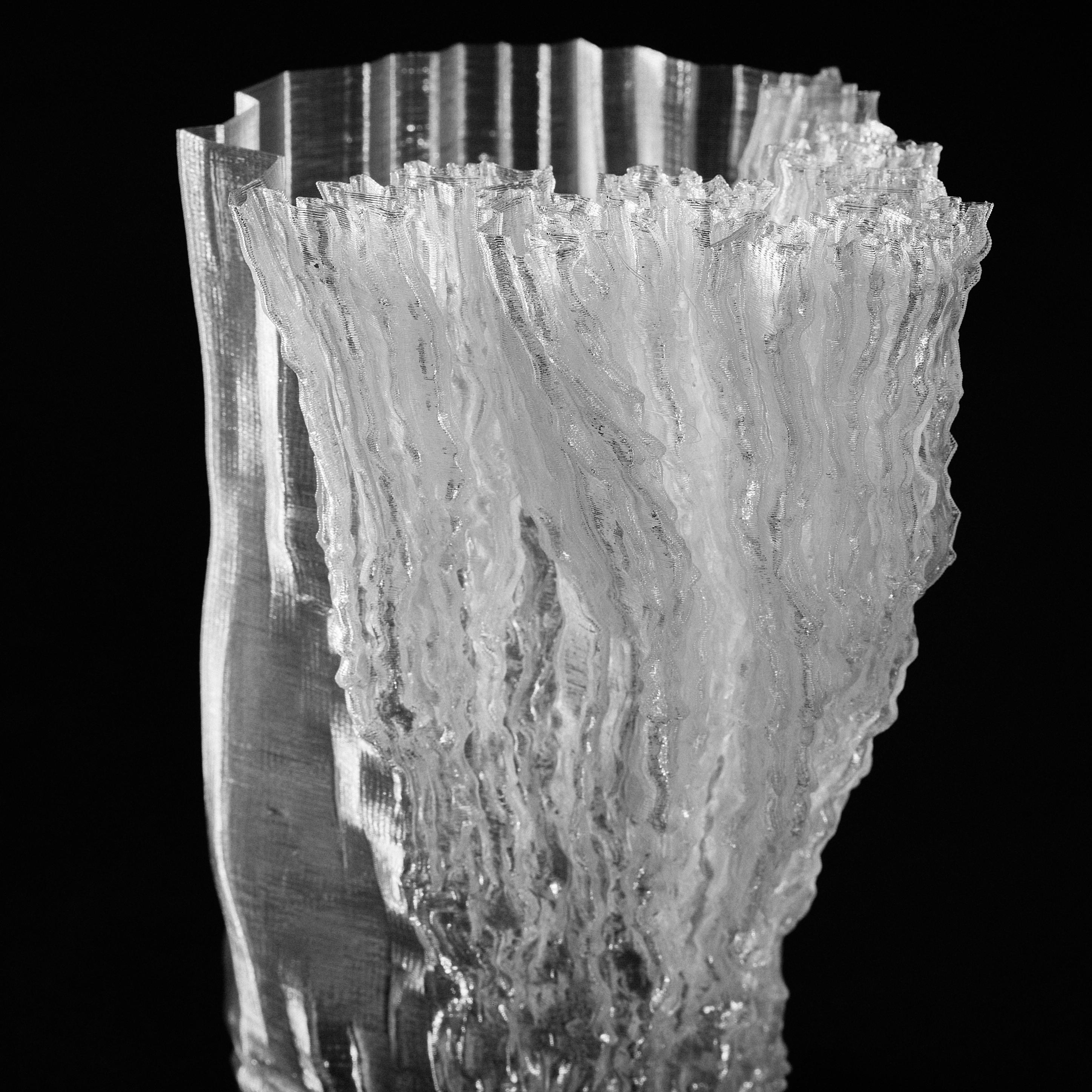} &
    \includegraphics[width=0.22\textwidth]{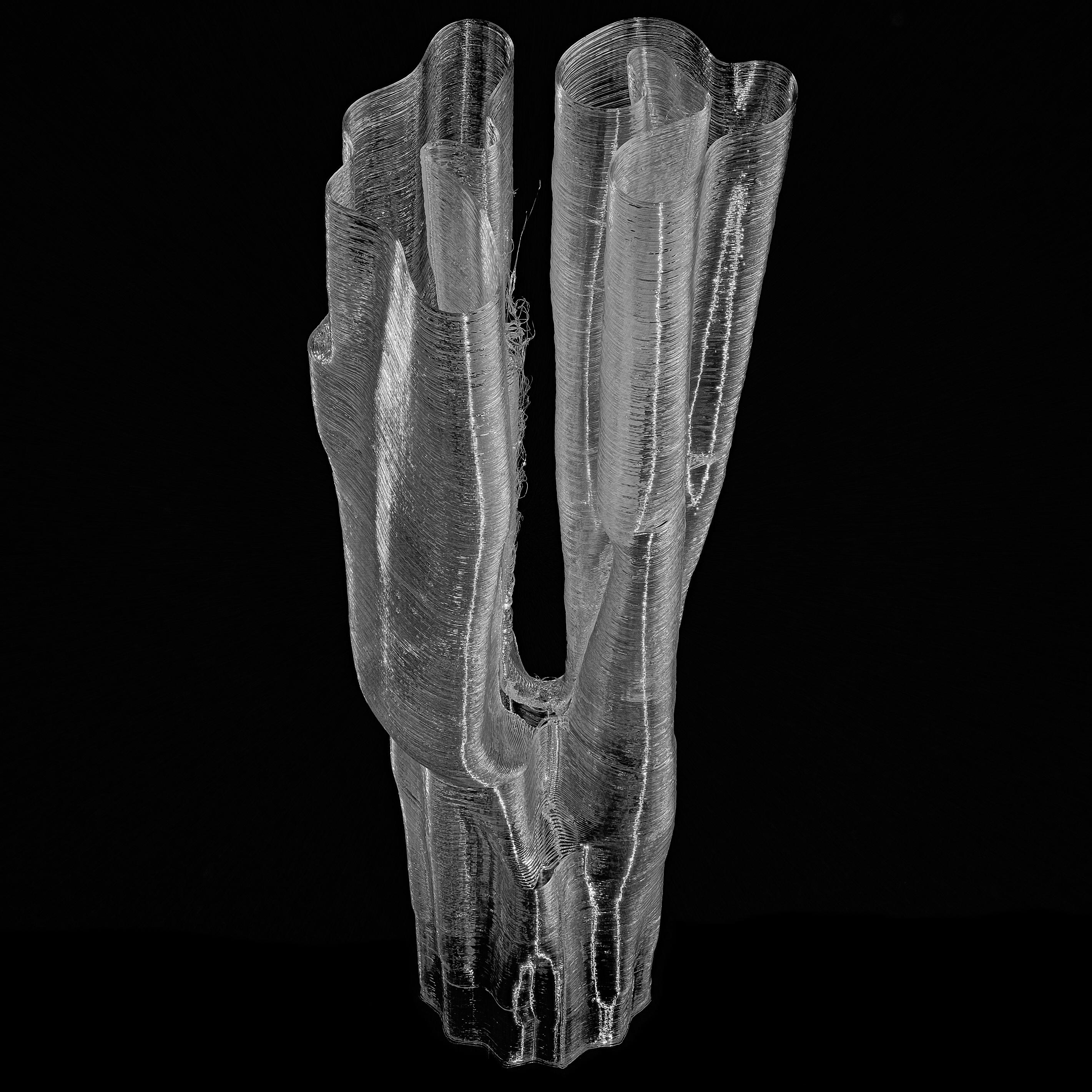} &
    \includegraphics[width=0.22\textwidth]{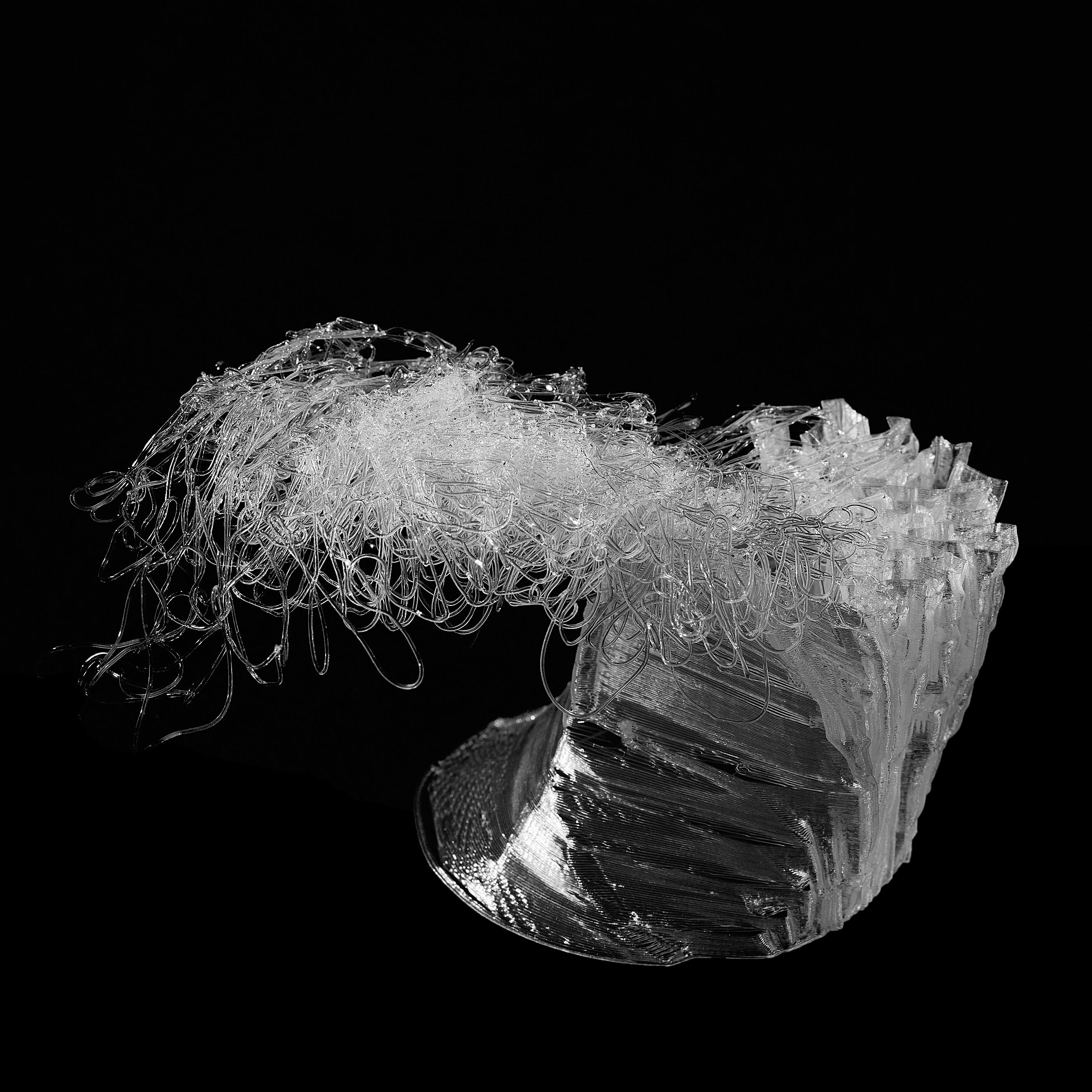}\\
    a & b & c & d
    \end{tabular}
    \caption{Example 3D printed objects generated during the development phase of our system, where a) is a detail of the object shown in Figure \ref{fig:heroForm} -- an object that developed as a single organism, b) and c) show fully printed objects that exhibit some aesthetic diversity, and d) shows an organism that failed to print.}
    \label{fig:preliminary_objects}
\end{figure*}

As shown in Figure \ref{fig:preliminary_objects}, successive layers are added on top of previous ones, turning a set of developing 2D polygons into a 3D object. The coherence of the developmental sequence ensures a degree of continuity between successive layers. The final object represents the developmental history of the 2D organism over a fixed number of time steps. The height of each layer was usually set to 0.2mm and run for 500 time steps during testing, producing an object up to 10cm in height. One limitation is the maximum height that the 3D printer is capable of printing, in the case of our experiments this was 20cm or 1,000 timesteps.

This method also has another important constraint if the form is to be successfully printed: each 2D layer cannot differ from the previous layer too significantly, otherwise the print will fail (known as \emph{layer coherence}). This is due to the extruded filament not being able to attach to the layer below and hence sagging or entangling the print head (see Figure \ref{fig:preliminary_objects}d), resulting in a failed print.

There are a number of possible ways to address this problem: a simple method would be to reduce the magnitude of the timestep $\Delta t$, effectively slowing down the development of the organism to reduce the possibility of unacceptable changes between layers. However this typically results in limited change overall and thus produces objects of little aesthetic interest.

A better alternative is to optimise the parameters that determine the organism's development so it grows in a way that produces aesthetically interesting but printable 3D forms. We discuss this strategy next.

\section{Evolutionary Form-finding}
\label{ss:es-form-finding}
We now describe the form-finding component of our system.
We use an evolutionary approach, based on the CMA-ES algorithm \cite{hansen2001completely}, to explore the generative capabilities of the simulation model described in the previous section. For this we introduce two distinct fitness measures that look at potentially conflicting characteristics of generated objects: \emph{structural consistency}, which attempts to measure the viability of the form to be successfully 3D printed, and \emph{formal complexity}, which in this case is used as a proxy of aesthetic appeal \cite{forsythe2011predicting,Johnson2019}.

We use evolution as a method of assisted exploration of the design space, rather than as a fully automated optimisation system. In any complex generative design system, there is no single best solution, rather a range of interesting or acceptable designs that are considered by the human designer. So our aim is to find the most promising regions of the design space effectively, but give the human designer an overall say in which designs are to be selected.

The use of two different fitness measures allows us to investigate the tension between the practical (structural consistency) and aesthetic (formal complexity) aspects of a designed artefact that designers traditionally face through their creative process. By dynamically changing fitness weights, properties of the object, such as regularity, fragmentation, texture and form tectonics emerge from the evolution of the system, rather than being determined through design operations, opening the possibility of unexpected and surprising design outcomes.

\subsection{Fitness functions}
\label{ss:fitness-functions}

As discussed, we calculate fitness for 3D printability ($P$) and aesthetic complexity ($C$). In our model these two criteria are often in tension: highly complex objects may not be printable and highly printable objects may not be interesting. Hence a possible design goal would be to generate objects that can be successfully printed yet are as complex as possible, understanding complexity as an approximation of aesthetic appeal.

\subsubsection{Structural Consistency}
\label{sss:structural-consistency}

We base the structural consistency (printability) measure of generated 3D objects on the constraints of Fused Deposition Modelling (FDM) (see Section \ref{ss:simulation-into-form}) -- a commonly available method for 3D printing \cite{Redwood2018}. We aim to print objects using a single strand of plastic filament with no additional supports as a way to minimise the amount of material waste from printing and to reduce post processing after the model is printed (e.g.~removal of support structures). However, for this to be feasible two conditions must be met: (i) the smallest diameter ($\diameter_{min}$) of the convex hull of all organisms cannot be smaller than a threshold $T$ (determined empirically) and (ii) each layer of material has to be physically supported by the layer that precedes it. 

To account for the diameter constraint we assign a diameter factor ($\diameter_f$) to each organism in an object based on the following conditions:

\begin{equation}
  \begin{gathered}
    \begin{aligned}
    \diameter_f & = \begin{cases}
        1 & \text{if } \diameter_{min} \geq T\\
        \diameter_{min} / T & \text{otherwise}
        \end{cases} \\
    \end{aligned}
  \end{gathered}
\end{equation}

where $\diameter_{min}$ is the smallest diameter of the convex hull of an organism, calculated using the \emph{rolling calipers} method \cite{Shamos1978computational}.

\begin{figure}
    \centering
    \includegraphics[width=0.45\textwidth]{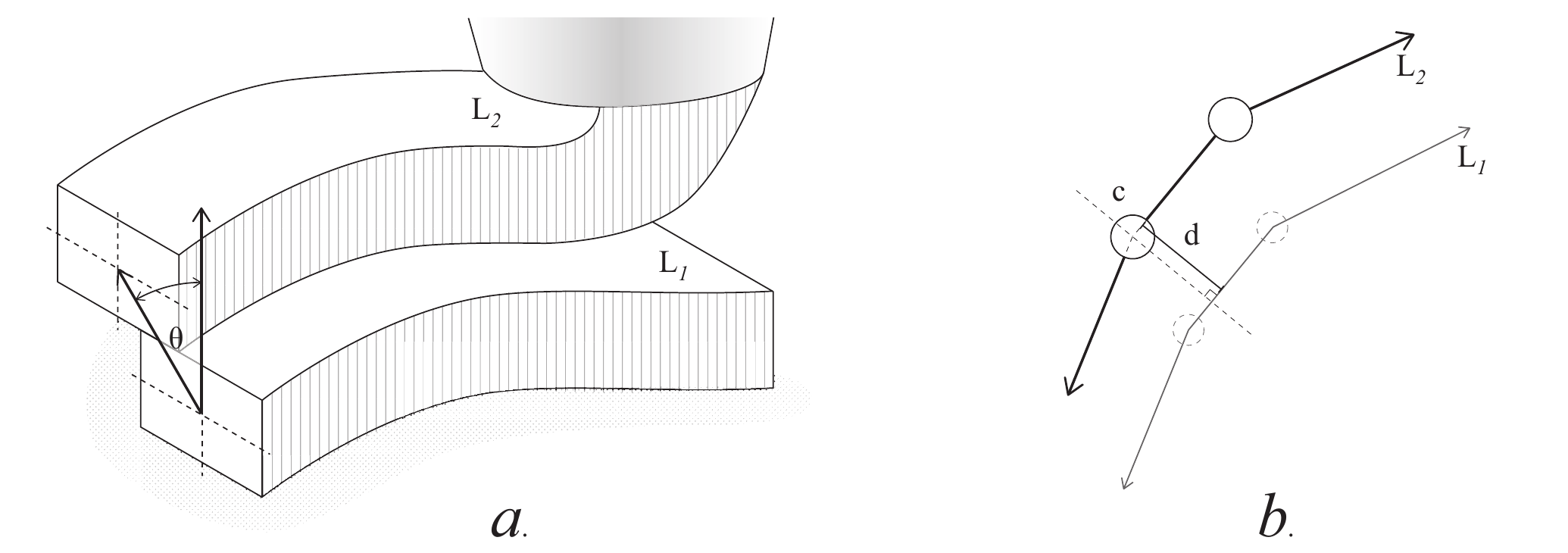}
    \caption{Layer support: (a) Axonometric projection showing stacked layers of filament ($L_1, L_2$), extruded from the print head, with open end faces. The angle of overhang ($\theta$) is measured between the vertical axis a layer and the line connecting the centre points of that layer and the one immediately above. This angle should not exceed $45^{\circ}$ (b) Top down projection of the centre axis of segments of filament.}
    \label{fig:support}
\end{figure}

In order to determine if a layer is supported we consider the recommendation of the 3D printer manufacturer, which indicates that for material to be supported the overhang between layers should not exceed 45 degrees (Figure \ref{fig:support}a). We use this principle to calculate a support score ($S$) for the vertices and midpoint of every edge in the object using the following equation:

\begin{equation}
  \begin{gathered}
    \begin{aligned}
    S & = \begin{cases}
        1 & \text{if } d \leq h\\
        2 - \frac{d}{h} & \text{if } h < d \leq 2 h\\
        0 & \text{otherwise}
        \end{cases} \\
    \end{aligned}
  \end{gathered}
\end{equation}

\noindent
where $d$ is the distance between the 2D projections of the vertex being assessed and the edge closest to it in the layer below (see Figure \ref{fig:support}b), and $h$ is the height of the layer.

We then use the support score of vertices and edge midpoints, in conjunction with $\diameter_f$ of the organism to calculate the ratio of the edge that is supported ($S_r$) as follows:

\begin{equation}
    S_r = \diameter_f \cdot ((S_{sp} + 2(S_{mp}) + S_{ep})/4)
\end{equation}

where $S_{sp}, S_{mp}, S_{ep}$ are the support scores for the start point, mid point and end point of the edge, respectively.

Finally, the overall printability for the colony, $P_{\mathcal{Y}} \in [0,1]$, is the normalised measure that the form can be successfully printed, calculated using the following equation:

\begin{equation}
    P_{\mathcal{Y}} = \frac{\sum_{i=1}^{n} S_{r_i} \cdot l_i}{\sum_{i=1}^{n} l_i}
\end{equation}

\noindent
where $n$ is the total number of edges in an object, $S_{r_i}$ is the support score of edge $i$ and $l_i$ is its length.

\subsubsection{Formal Complexity}
\label{sss:formal-complexity}
 We employed two separate metrics to evaluate the complexity of a candidate form: \emph{convexity} and the quartile \emph{coefficient of dispersion} of angles between consecutive edges. Each measure is calculated for every organism in the colony at each time step.

Convexity, $X$, is calculated as the ratio between the perimeter of the convex hull of a shape and its total perimeter, i.e.
\begin{equation}
X = \frac{L(H(\mathcal{O}_t))}{L(\mathcal{O}_t)},
\end{equation}

\noindent
where $L : \texttt{Organism} \to \mathbb{R} = \sum^{\mathcal{O}_t}|e_{i}|$ is a function that returns the perimeter length of $\mathcal{O}$ at time $t$, and $H : \texttt{Organism} \to \mathbb{R}$ a function that returns the perimeter length of the convex hull of $\mathcal{O}$. A completely convex shape will yield a convexity score of 1. Conversely, a shape with a `rougher' surface will yield a lower score.

The quartile coefficient of dispersion of angles is calculated as follows:

\begin{equation}
    D = \frac{Q_{3} - Q_1}{Q_3 + Q_1}
\end{equation}
\noindent
where $Q_1$ and $Q_3$ are the first and third quartiles of the sorted array of all angles between consecutive edges  $(e_i, e_{i+1})$ in the 3D shape. A low dispersion score indicates a lack of diversity of angles, which translates into a regular shape.

To calculate overall formal complexity of a colony, $C_{\mathcal{Y}}$, we use a weighted sum: $C_{\mathcal{Y}} = aX_{\mathcal{Y}} + (1-a)D_{\mathcal{Y}}$, where $a$ is the normalised relative weighting.
Similarly, overall fitness is the weighted sum of $P_{\mathcal{Y}}$ and $C_{\mathcal{Y}}$. We experimented with different weights and changing weights during an evolutionary run, described in Section \ref{s:experiments}.

\subsection{Evolution strategy} 
\label{ss:evo_strategy}

We used an implementation of the covariance matrix adaptation evolution strategy (CMA-ES) \cite{hansen2001completely}, because it is well suited to work with low-dimensional floating point vectors as genomes, directly matching the genetic encoding of our system. Additionally, its adaptive capabilities enable us to purposefully explore a fitness landscape that, as revealed by our preliminary studies, is far from unimodal. Lastly, it allows us to work with variable weight fitness functions during a single evolutionary run.

\subsection{Model Design Space}
\label{ss:model_design_space}
To better understand the design space of our form generation model we generated a number of individual forms to examine the range and diversity of possible forms. Fixed aspects built into the model constrain the overall size and basic structure of any possible form. These constraints include the build area of the 3D printer and the underlying stacked 2D, line-based geometry used to generate toolpah movements. Nevertheless the range of possible forms possible is quite diverse. A dataset of 2,500 generated forms is available for the interested reader to examine \cite{McCormack2021_DLADataset}. We generated 9,950 uniform random initialised forms and measured their $P_{\mathcal{Y}}$ and $C_{\mathcal{Y}}$ fitness values. The mean $P_{\mathcal{Y}}$ was 0.946 with $\sigma = 0.175$. Mean  $C_{\mathcal{Y}}$ was 0.105 with $\sigma = 0.1$. This indicates that the majority of forms are highly printable but aesthetically uninteresting. Hence we turn to evolution to find highly complex yet printable forms.

\section{Experiment Settings and Results}
\label{s:experiments}
In this section we present a series of experiments running the CMA-ES
on the generative system to evolve suitable designs. The two approaches we explored consider a) evolving generated colonies using different fitness weights, and b) altering weights at different points in an evolutionary run.

\subsection{Initial parameters for the generation of organisms}
The evolution of 3D printable objects using our generative developmental system is based on the progressive transformation of the five alleles in the genome of an organism ($\eta, \nu, \varepsilon_{max}, k, \rho$). We defined the remaining parameters of the system as follows:
 
\subsubsection{Environment}
\label{sss:environment}
The environment is defined as a 2D plane that corresponds to 3D printer's build plate dimensions. We used a resolution of $600 \times 600$ units (0.35mm/unit) giving sufficient detail for intricate shapes and motifs balanced with reasonable processing times. The placement and diffusion of nutrients in the environment is discreetised into $15 \times 15$ tiles, with number of sources $n_S = 5, \alpha_{S_i} = 0.1, i \in [1..5]$ and $\beta_{S_i} = \textit{URND}(5,10)$, where $\textit{URND}(a,b) \to \mathbb{R}$ returns a uniformly distributed pseudo-random number in the range $[a,b)$ (refer Section \ref{ss:food}).

\subsubsection{Colonies}

The colony $\mathcal{Y}$ is the unit of selection in the evolutionary component of our system, i.e.~%
we evolve populations of colonies. Each colony $\mathcal{Y}_n$ generated by the evolutionary system is initialised with a single circular organism with its centroid at the centre of the build plate, a radius of 100 units and initial energy, $\varepsilon_{init} = 5$, for each cell. The number of initial cells in the organism is set to  $2\pi r / 2\varepsilon$, where $r$ is the initial organism radius (100). The repulsion radius, $r_r$ for the cells is set to 3 and the cost of living $c_1 = 10^{-3}$. Additionally, the diffusion rate is set to $\chi = 0.1$ and loss of energy per unit distance $\omega = 10 \times 5^{-2}$   (refer Section \ref{ss:energy}).

\subsection{Parameters for the evolutionary processes}


A hyperparamerter search was conducted to determine the CMA-ES values for $\lambda$, $\mu$ and the standard deviation for initial sampling $\sigma_{i}$, with $\lambda = 40$, $\mu = 2$ and $\sigma_{i} = 0.1$ found to offer the best trade-off between search efficiency and compute resources.
The remaining parameters, maximum iterations $g$ and step size (learning rate) $\sigma$ used different values for each experiment. For the initial experiments, $g = 250$ and $\sigma = 10^{-3}$. 

\subsection{Experiments and Results} 
\label{ss:results}

\begin{figure*}[htbp]
    \centering
    \begin{tabular}{cc}
    \includegraphics[width=0.47\textwidth]{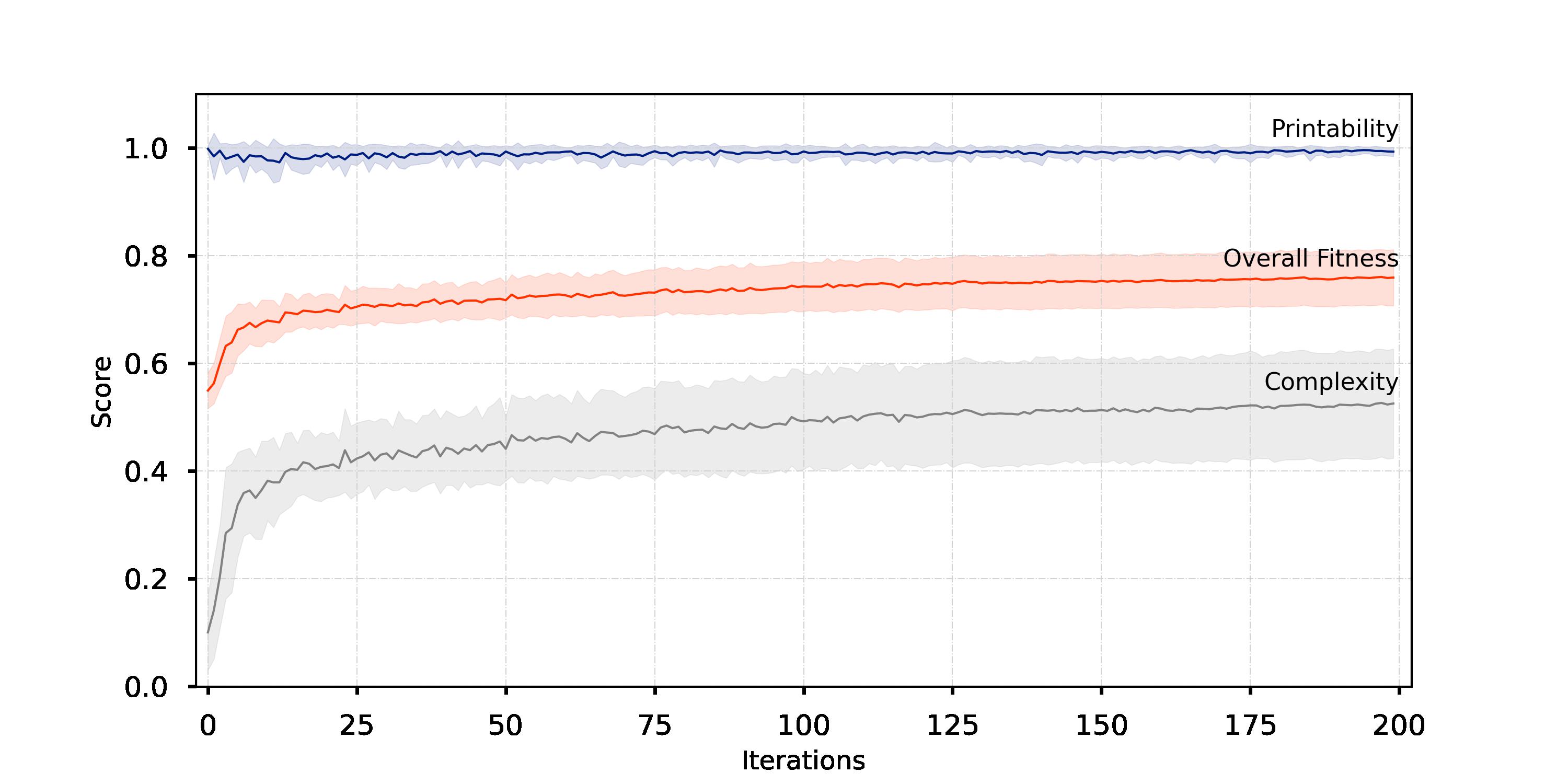} &     \includegraphics[width=0.47\textwidth]{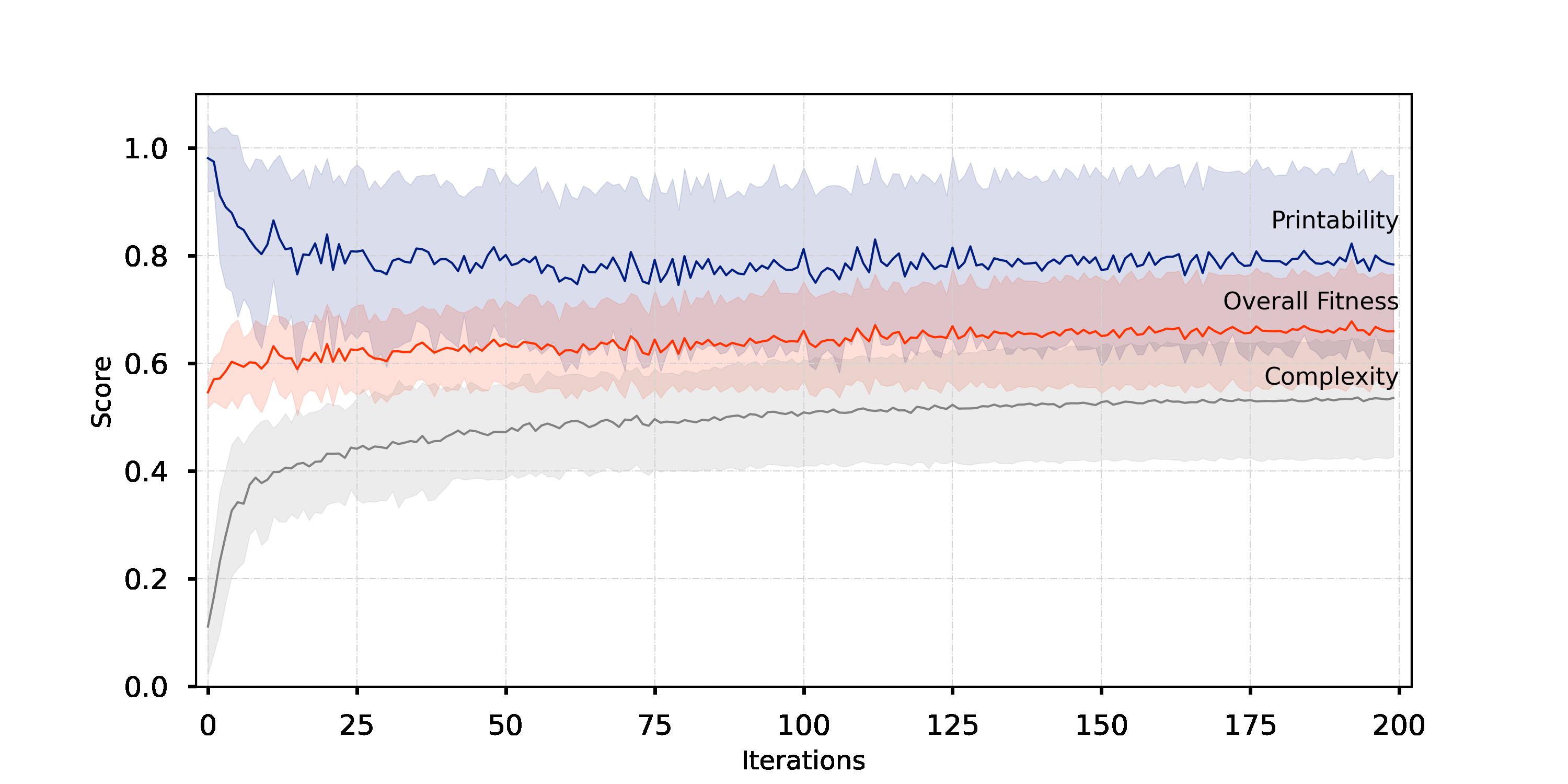} \\
    a & b
    \end{tabular}
    \caption{Evolutionary runs showing the mean printability ($P_{\mathcal{Y}}$), complexity ($C_{\mathcal{Y}}$) and overall fitness for the best individual at every generation averaged over 30 runs. The shaded components show the variance. In a) the system was set to optimise $P_{\mathcal{Y}}$ and $C_{\mathcal{Y}}$ equally, and in b) only $C_{\mathcal{Y}}$.}
    \label{fig:means}
\end{figure*}

Our first experiments tested the ability of the system to find designs using a weighted sum of the fitness measures $P_{\mathcal{Y}}$ and $C_{\mathcal{Y}}$. Figure \ref{fig:means} shows the results averaged over 30 runs, with each run using a new random seed for the environment (Section \ref{sss:environment}) -- designed to eliminate environmental dependencies when evaluating the overall ES performance. Figure \ref{fig:means}a, shows the results using equal weighting. As can be seen, the printability quickly optimises while there is a modest increase in complexity. Visually examining the fittest individuals at each generation revealed that once the system had found a highly printable form, it was difficult to escape the basic structure of that form to find more complex objects that remain as printable. 

We experimented with increasing the weighting of $C_{\mathcal{Y}}$, and ultimately found that maximising the weighting gave the most interesting design results (Figure \ref{fig:means}b). In this case, evolution was better able to optimise complexity without the additional constraint of printability, confirming what is often intuitive to experienced human designers: that practical constraints often limit the overall complexity a form can achieve.

\begin{figure*}[htbp]
    \centering
    \includegraphics[width = \textwidth]{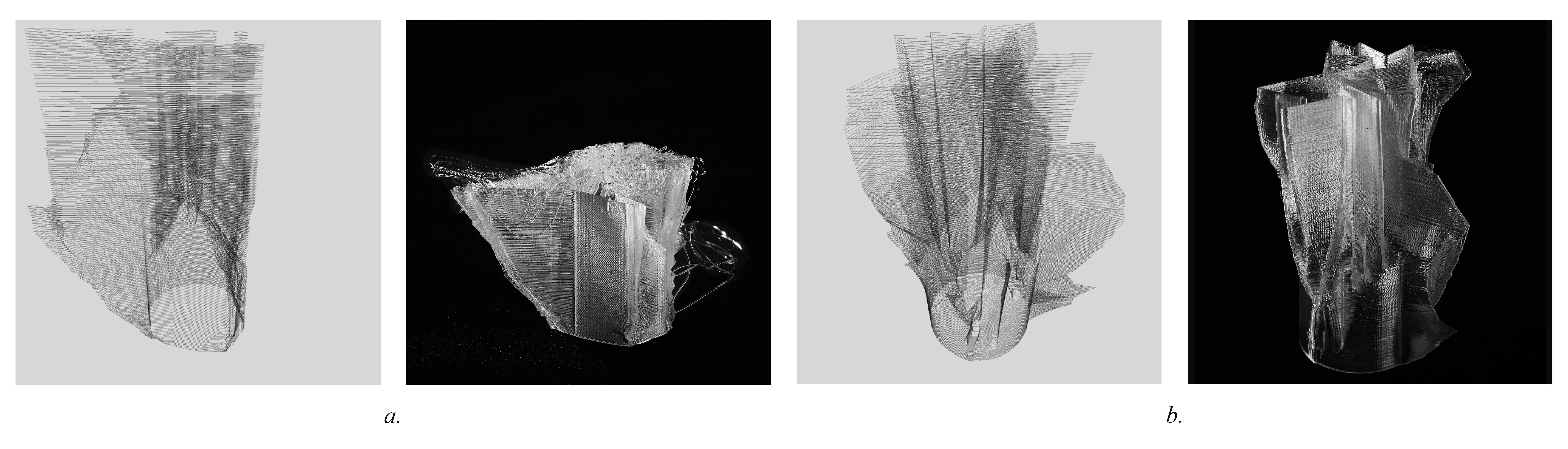}
    \caption{Side by side comparison of digital rendering and 3D printed versions of two generated forms. For (a) $C_{\mathcal{Y}} = 0.417$ and $P_{\mathcal{Y}} = 0.695$, resulting in a failed print. For (b) $C_{\mathcal{Y}} = 0.516$ and $P_{\mathcal{Y}} = 1.0$, giving a fully printable form.}
    \label{fig:objects}
\end{figure*}

Figure \ref{fig:objects} shows example outcomes when evolving colonies optimising for complexity and illustrates the consequences of a low printability score. The two images on the left (a) show the digital rendering and 3D printed versions of a form with $C_{\mathcal{Y}} = 0.417$ and $P_{\mathcal{Y}} = 0.695$, resulting in a failed print. The images on the right (b) show an object with higher complexity ($C_{\mathcal{Y}} = 0.516$) and printability ($P_{\mathcal{Y}} = 1.0$), which prints successfully.

\begin{figure}[htbp]
    \centering
    \includegraphics[width=0.45\textwidth]{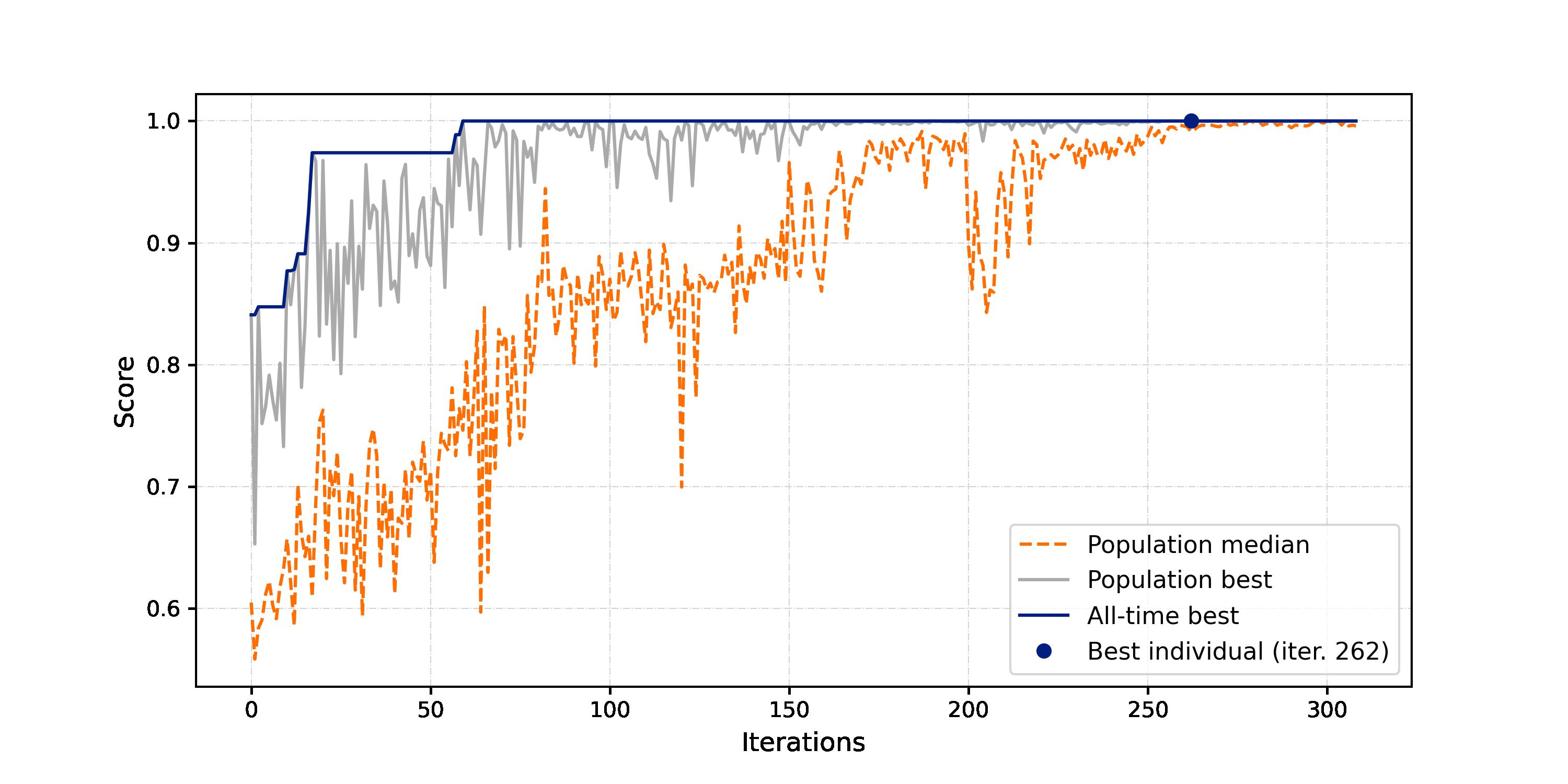}
    \caption{Evolutionary run initialised with an individual selected for its complexity and aesthetic interest. This individual is used as reference point to sample the initial population of an evolutionary process aimed at maximising $P_{\mathcal{Y}}$.}
    \label{fig:printability_optimisation}
\end{figure}

\begin{figure}[htbp]
    \centering
    \begin{tabular}{cc}
    \includegraphics[width=0.22\textwidth]{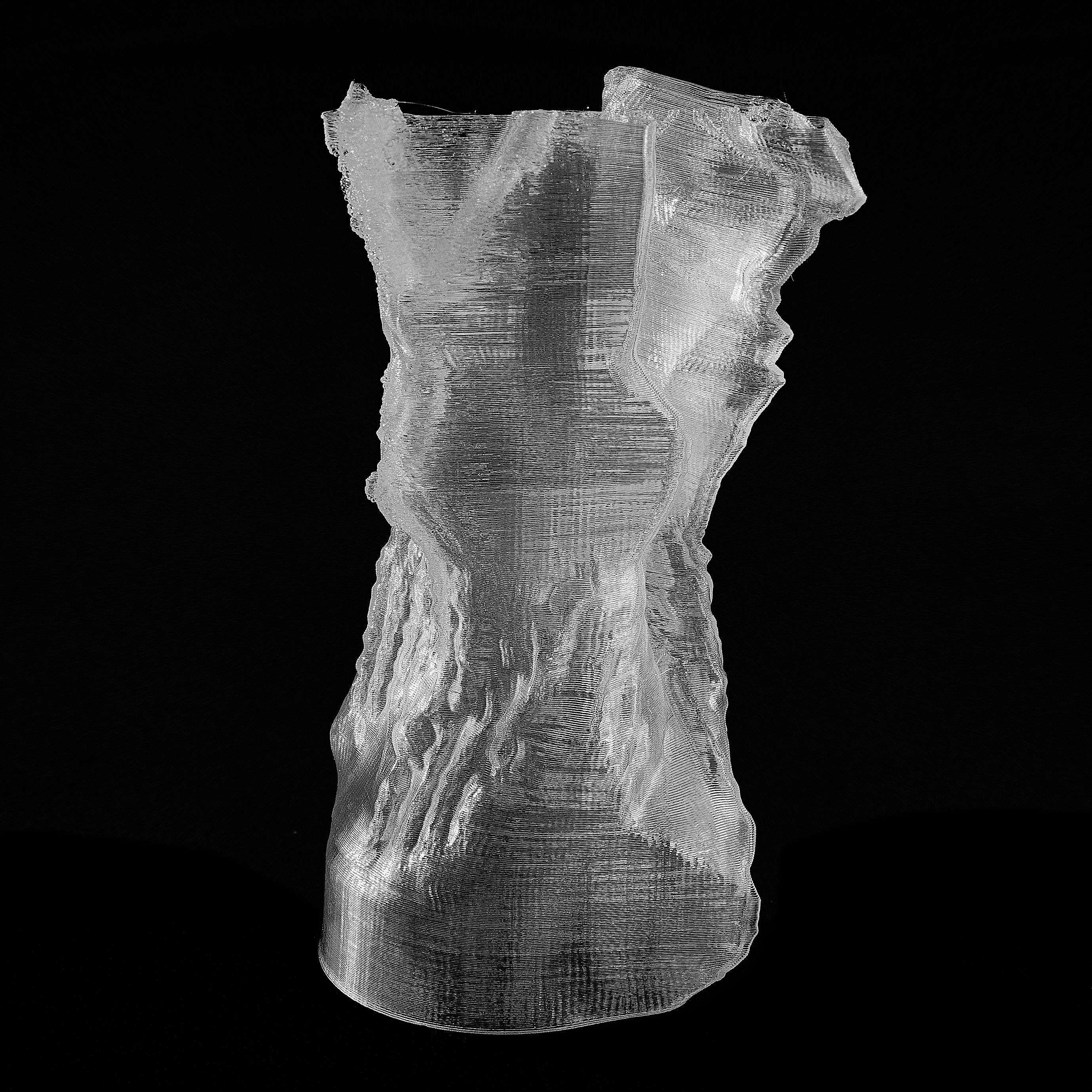} &
    \includegraphics[width=0.22\textwidth]{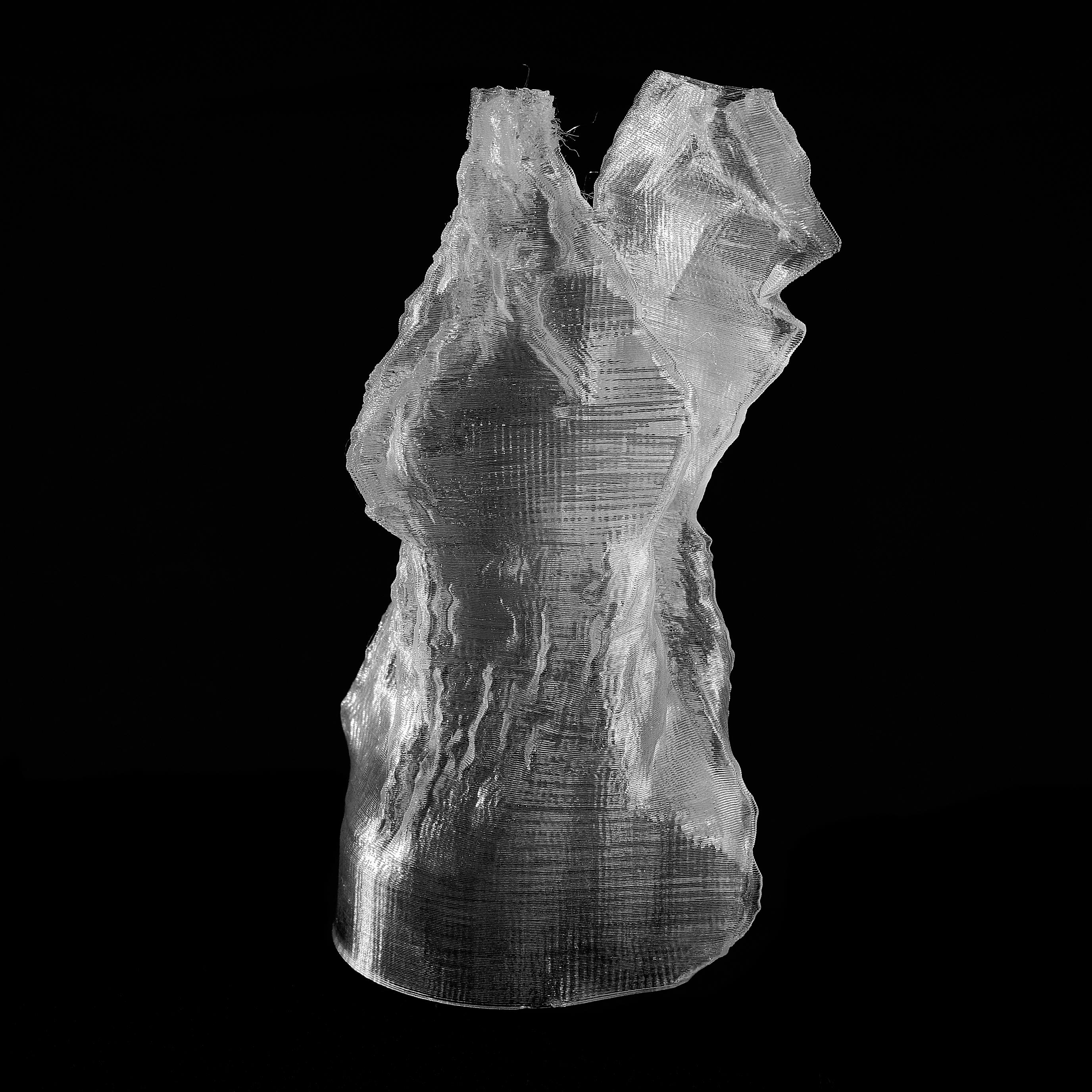}\\
    a & b
    \end{tabular}
    \caption{Initial (a) and final form (b) evolved to optimise $P_{\mathcal{Y}}$.}
    \label{fig:objects_exp2}
\end{figure}

In the second set of experiments, we attempted to improve the printability of specific objects generated in the previous stage while preserving their physical complexity. The rationale for this is to ``tame'' an interesting design back to just being printable while trying to preserve its complexity. To do so we selected visually interesting individuals (high $C_{\mathcal{Y}}$) with low $P_{\mathcal{Y}}$ measures from the first set of evolutionary runs, and used them as reference individuals when sampling the initial population. The seed for the environment was carried over from the seed individual's initial evolution, to ensure an identical form. For these runs the number of maximum iterations was set to $g = 400$, the initial standard deviation to $\sigma_i = 10^{-4}$  and the learning rate to $\sigma = 10^{-6}$, favouring local search around the initial form. Figure \ref{fig:printability_optimisation} shows how CMA-ES managed to improve $P_{\mathcal{Y}}$. Figure \ref{fig:objects_exp2} shows example initial and final objects. It is possible to observe that while both objects are different, they maintain similarities in overall shape and complexity -- especially in the lower section. A closer inspection of the finished products reveals deficiencies in the quality of the initial print, that are corrected in the final outcome.

\section{Analysis and Discussion}
\label{s:analysis}

The results demonstrate that our generative developmental system can be effective for producing objects with a diverse range of physical characteristics and unique aesthetic properties.       

\subsection{Fitness Measures}
\label{ss:fitness-measures}
The two new fitness measures we introduced (Section \ref{ss:fitness-functions}) were designed to evaluate the opposing tensions of structural consistency (printability) and formal complexity (aesthetics). We empirically tested and tuned these measures on hundreds of simulations and many dozens of physical prints to arrive with the final formulations presented. 

Using a continuous measure of printability allows the designer to decide how much they are willing to risk a print failing in order to create a specific aesthetic form. Empirically testing our measure on dozens of actual 3D prints showed it to be a reliable indicator of how well a form will print, with a measure of 1 being effectively `perfect' (always printable without flaws) and  values $> 0.9$ having very slight, occasional flaws but are always fully printable. Values below 0.8 have noticeable flaws but are generally still fully printable, whereas those below 0.7 will often fail to fully print at all (Figure \ref{fig:objects}a). Having a reliable measure such as this is important. Complex 3D prints can take many hours (or even days in extreme cases) to print and require significant quantities of print material. Both time and material are wasted on a failed print. Knowing that a form will successfully print eliminates this problem.

Our complexity measure was developed as a quantifiable proxy for visual aesthetics. This intuition is based on past findings that associate complexity with aesthetics \cite{forsythe2011predicting,Johnson2019}. As we discussed in Section \ref{ss:evolutionaryMethods}, human aesthetic judgement depends on many factors beyond the visual or structural appearance of objects. Hence it is difficult to objectively evaluate the effectiveness of our measure in absolute terms, beyond our observations as practising designers that objects with higher complexity measures seem more interesting to us.

However, an additional study using a dataset of 2,500 forms \cite{McCormack2021_DLADataset} confirmed that our formal complexity measure has a high correlation to visual complexity. We evaluated the Pearson correlation between $C_{\mathcal{Y}}$ and the 2D image visual complexity measure described in \cite{Lakhal2020}, achieving a correlation of 0.774 with $p < 10^{-2}$ \cite{McCormack2021Enigma}. Hence we believe we can claim our measure is comparable with other leading measures of visual complexity for the forms generated by our system.

\subsection{Evolution}
\label{ss:evolution}

Adjusting the weights of the components of the fitness function over evolutionary runs and the learning rate to find and refine complex objects proved to be viable technique for the exploration of the useful design space produced by the generative system.

By observing the evolution of the fittest individual in the population, when the system is set to optimise for complexity, the CMA-ES approach is capable of finding fitter individuals over time. However, without taking into account the printability, some complex designs will fail to successfully print. Reducing the learning rate favours localised search, in which characteristics of the objects being generated are preserved, but their printability increases significantly.

Comparing the evolutionary trajectories of the populations evolved with differently weighted fitness functions, two main observations can be made. Firstly, it is not possible to draw an absolute correlation between complexity and printability, as can be seen in Figure \ref{fig:means}(b). However, Figure \ref{fig:means}(a) shows that by weighting printability and complexity equally, the search for more complex objects is less effective. This is most likely due to the nature of the fitness landscape, which has a large number of forms with high $P_{\mathcal{Y}}$ but medium to low $C_{\mathcal{Y}}$ scores.

In light of the results obtained, the flexibility shown by the generative system, and the capability of the CMA-ES strategy to improve the fitness of generated objects over time, we foresee two main avenues to continue our exploration. First, the revision and adjustment of the method that measures complexity seems to be paramount, as some formal aspects of the generated objects that have a significant impact on printability, as well as on their perceived complexity, such as the repeated division of organisms -- which results in branching 3D objects -- are not sufficiently reflected in complexity scores. 

Second, using the adaptive capabilities of CMA-ES, it should be possible to dynamically vary fitness and learning weights as a way to explore the design space interactively. Compute times to generate populations remains the barrier to doing this in real time, so greater efficiencies in our generative model need to be implemented.

 Given this could be considered a multi-objective optimisation problem, it is reasonable to ask why MOEA methods such as NSGA-II \cite{Deb2002} were not used in our system. While the MOEA approach is more flexible in its ability to allow wider  candidate solutions with diverse features to be evolved at the same time, crossover is capable of capturing diversity more consistently when performed on larger (higher dimension) feature vectors. Conversely the CMA-ES approach works well with low-dimensional genome vectors, and requires a much smaller population size, which leads to faster execution, especially when it finds a good solution. Given the significant processing time to grow a colony into a final form, we opted for this efficiency, hoping for more interactive explorations of the design space through many short evolutionary runs. Future work could certainly look into the use of MOEA as an alternative to CMA-ES.

\section{Conclusion}
\label{s:conclusion}

In this paper we have described a system that explores a design space to find 3D forms with high practical and aesthetic fitness. The system was developed in two stages: First, a generative developmental system was devised that is capable of generating a diverse range of 3D forms via the manipulation of a low-dimensional parameter vector; and second, an evolutionary algorithm that incorporates printability and aesthetic fitness measures was used to explore the design space produced by the generative system.

We tested the system by evolving populations of generated 3D objects suitable for 3D printing. The primary goal was to find candidates that were both printable and aesthetically complex. However, this exploration also provided insight into the characteristics of the design space itself, via the comparative analysis of the generated objects.

The results show that the evolution strategy used provides a reliable search mechanism for the exploration of design space, as long as the fitness functions accurately represent the design objectives being pursued. Being able to dynamically adjust the weights between practical (printability) and aesthetic (visual complexity) goals as the evolution progresses opens the possibility of a designer-guided exploration of the design space. In contrast to algorithms such as the IGA, we automate both aesthetic and practical optimisation goals, removing the designer from selecting the fittest individual at every generation. Instead, after evolving for specific criteria, the designer can then shift the weighting of fitness criteria and learning rate for selected high fitness individuals, then using the evolutionary system to maintain one evolved trait (e.g.~Complexity) while improving the other (Printability). This form of guided evolutionary search allows the designer to focus on selection of individuals at specific points in the overall process, leaving evolution to do the optimisation via formalised fitness measures.


\section*{Acknowledgements}

This research was supported by an Australian Research Council grant FT1701\-00033. 

\ifCLASSOPTIONcaptionsoff
  \newpage
\fi





\bibliographystyle{IEEEtran}
\bibliography{main}

\begin{thebibliography}{10}
\providecommand{\url}[1]{#1}
\csname url@rmstyle\endcsname
\providecommand{\newblock}{\relax}
\providecommand{\bibinfo}[2]{#2}
\providecommand\BIBentrySTDinterwordspacing{\spaceskip=0pt\relax}
\providecommand\BIBentryALTinterwordstretchfactor{4}
\providecommand\BIBentryALTinterwordspacing{\spaceskip=\fontdimen2\font plus
\BIBentryALTinterwordstretchfactor\fontdimen3\font minus
  \fontdimen4\font\relax}
\providecommand\BIBforeignlanguage[2]{{%
\expandafter\ifx\csname l@#1\endcsname\relax
\typeout{** WARNING: IEEEtran.bst: No hyphenation pattern has been}%
\typeout{** loaded for the language `#1'. Using the pattern for}%
\typeout{** the default language instead.}%
\else
\language=\csname l@#1\endcsname
\fi
#2}}

\bibitem{hillier1989social}
B.~Hillier and J.~Hanson, \emph{The social logic of space}.\hskip 1em plus
  0.5em minus 0.4em\relax Cambridge university press, 1989.

\bibitem{lawson1990designers}
B.~Lawson, \emph{How designers think: The design process demystified},
  2nd~ed.\hskip 1em plus 0.5em minus 0.4em\relax London: Routledge, 1990.

\bibitem{Mitchell1977}
W.~J. Mitchell, \emph{Computer-aided architectural design}.\hskip 1em plus
  0.5em minus 0.4em\relax New York: John Wiley \& Sons, Inc., 1977.

\bibitem{Adamatzky2016}
A.~Adamatzky and G.~J. Martinez, Eds., \emph{Designing Beauty: The Art of
  Cellular Automata}, ser. Emergence, Complexity and Computation.\hskip 1em
  plus 0.5em minus 0.4em\relax Springer International Publishing, 2016.

\bibitem{McCormack2005c}
J.~McCormack, ``A developmental model for generative media,'' in \emph{Advances
  in Artificial Life (8th European Conference, ECAL 2005)}, M.~Capcarrere,
  A.~A. Freitas, P.~J. Bentley, C.~G. Johnson, and J.~Timmis, Eds.\hskip 1em
  plus 0.5em minus 0.4em\relax Berlin; Heidelberg: Springer-Verlag, 2005, vol.
  LNAI 3630, pp. 88--97.

\bibitem{GreenfieldM09}
G.~Greenfield and P.~Machado, ``Simulating artist and critic dynamics - an
  agent-based application of an evolutionary art system,'' in \emph{IJCCI},
  A.~D. Correia, A.~C. Rosa, and K.~Madani, Eds.\hskip 1em plus 0.5em minus
  0.4em\relax INSTICC Press, 2009, pp. 190--197.

\bibitem{Jones2015}
\BIBentryALTinterwordspacing
J.~Jones, \emph{From Pattern Formation to Material Computation}, ser.
  Emergence, Complexity and Computation.\hskip 1em plus 0.5em minus 0.4em\relax
  Springer International Publishing, 2015. [Online]. Available:
  \url{http://dx.doi.org/10.1007/978-3-319-16823-4}
\BIBentrySTDinterwordspacing

\bibitem{Bentley1999}
P.~J. Bentley, \emph{Evolutionary design by computers}.\hskip 1em plus 0.5em
  minus 0.4em\relax San Francisco, Calif.: Morgan Kaufmann Publishers, 1999.

\bibitem{BentleyCorne2002}
P.~J. Bentley and D.~W. Corne, Eds., \emph{Creative Evolutionary
  Systems}.\hskip 1em plus 0.5em minus 0.4em\relax London: Academic Press,
  2002.

\bibitem{lynn1999animate}
G.~Lynn and T.~Kelly, \emph{Animate form}.\hskip 1em plus 0.5em minus
  0.4em\relax Princeton Architectural Press New York, 1999.

\bibitem{Porter2010}
B.~Porter and J.~McCormack, ``Developmental modelling with {SDS},''
  \emph{Computers \& Graphics}, vol.~34, no.~4, pp. 294--303, 2010.

\bibitem{dino2012creative}
I.~Dino, ``Creative design exploration by parametric generative systems in
  architecture,'' \emph{METU Journal of Faculty of Architecture}, vol.~29,
  no.~1, pp. 207--224, 2012.

\bibitem{munk2015topology}
D.~J. Munk, G.~A. Vio, and G.~P. Steven, ``Topology and shape optimization
  methods using evolutionary algorithms: a review,'' \emph{Structural and
  Multidisciplinary Optimization}, vol.~52, no.~3, pp. 613--631, 2015.

\bibitem{waibel2019building}
C.~Waibel, T.~Wortmann, R.~Evins, and J.~Carmeliet, ``Building energy
  optimization: An extensive benchmark of global search algorithms,''
  \emph{Energy and Buildings}, vol. 187, pp. 218--240, 2019.

\bibitem{Redwood2018}
B.~Harwood, F.~Sch\"{o}ffer, and B.~Garret, \emph{The 3D Printing
  Handbook}.\hskip 1em plus 0.5em minus 0.4em\relax The Netherlands: 3D Hubs
  B.V., 2018.

\bibitem{Steuben2016}
J.~C. Steuben, A.~P. Iliopoulos, and J.~G. Michopoulos, ``Implicit slicing for
  functionally tailored additive manufacturing,'' \emph{Computer-Aided Design},
  vol.~77, pp. 107--119, 2016.

\bibitem{mccormack2014ten}
J.~McCormack, O.~Bown, A.~Dorin, J.~McCabe, G.~Monro, and M.~Whitelaw, ``Ten
  questions concerning generative computer art,'' \emph{Leonardo}, vol.~47,
  no.~2, pp. 135--141, 2014.

\bibitem{Whitelaw2005}
M.~Whitelaw, ``System stories and model worlds: A critical approach to
  generative art,'' in \emph{Readme 100: temporary software art factory}.\hskip
  1em plus 0.5em minus 0.4em\relax Norderstedt: Herstellung und Verlag, 2005,
  pp. 135--156.

\bibitem{McCormack2004b}
\BIBentryALTinterwordspacing
J.~McCormack, A.~Dorin, and T.~Innocent, ``Generative design: A paradigm for
  design research,'' in \emph{Futureground}, J.~Redmond, D.~Durling, and
  A.~de~Bono, Eds.\hskip 1em plus 0.5em minus 0.4em\relax Melbourne, Australia:
  Design Research Society, 2004. [Online]. Available:
  \url{https://dl.designresearchsociety.org/drs-conference-papers/drs2004/researchpapers/171/}
\BIBentrySTDinterwordspacing

\bibitem{BodenEdmonds2009}
M.~A. Boden and E.~A. Edmonds, ``What is generative art?'' \emph{Digital
  Creativity}, vol.~20, no. 1 \& 2, pp. 21--46, March 2009.

\bibitem{reas2010}
\BIBentryALTinterwordspacing
C.~Reas, C.~McWilliams, and LUST, \emph{Form+Code in Design, Art, and
  Architecture}.\hskip 1em plus 0.5em minus 0.4em\relax Princeton Architectural
  Press, Inc., 2010. [Online]. Available: \url{http://formandcode.com/}
\BIBentrySTDinterwordspacing

\bibitem{Dorin2012}
\BIBentryALTinterwordspacing
A.~Dorin, J.~McCabe, J.~McCormack, G.~Monro, and M.~Whitelaw, ``A framework for
  understanding generative art,'' \emph{Digital Creativity}, vol.~23, no. 3-4,
  pp. 239--259, 2012. [Online]. Available:
  \url{https://doi.org/10.1080/14626268.2012.709940}
\BIBentrySTDinterwordspacing

\bibitem{Bohnacker2012}
H.~Bohnacker, B.~Gro{\ss}, J.~Laub, and C.~Lazzeroni, \emph{Generative Design:
  Visualize, Program, and Create with Processing}, {E}nglish~ed.\hskip 1em plus
  0.5em minus 0.4em\relax Princeton Architectural Press, 2012.

\bibitem{Stevens1974}
P.~S. Stevens, \emph{Patterns in nature}.\hskip 1em plus 0.5em minus
  0.4em\relax Boston, Mass.: Little Brown, 1974.

\bibitem{Alexander1964}
C.~Alexander, \emph{Notes on the synthesis of form}.\hskip 1em plus 0.5em minus
  0.4em\relax Cambridge, Mass: Harvard University Press, 1964.

\bibitem{alexander1977pattern}
------, \emph{A pattern language: towns, buildings, construction}.\hskip 1em
  plus 0.5em minus 0.4em\relax Oxford university press, 1977.

\bibitem{Stiny1975}
G.~Stiny, \emph{Pictorial and formal aspects of shape and shape grammars}, ser.
  ISR, Interdisciplinary systems research;.\hskip 1em plus 0.5em minus
  0.4em\relax Basel; Stuttgart: Birkh{\"a}user, 1975, no. xv, 399.

\bibitem{whitelaw2015accretor}
M.~Whitelaw, ``Accretor: Generative materiality in the work of {Driessens and
  Verstappen},'' \emph{Artificial life}, vol.~21, no.~3, pp. 307--312, 2015.

\bibitem{lomas2014cellular}
A.~Lomas, ``Cellular forms: an artistic exploration of morphogenesis,'' in
  \emph{AISB}.\hskip 1em plus 0.5em minus 0.4em\relax ACM, 2014.

\bibitem{Lomas2019}
\BIBentryALTinterwordspacing
------, ``Morphogenetic vase forms,'' \emph{Artificial Life Conference
  Proceedings}, no.~31, pp. 523--530, 2019. [Online]. Available:
  \url{https://www.mitpressjournals.org/doi/abs/10.1162/isal_a_00215}
\BIBentrySTDinterwordspacing

\bibitem{Sims1991}
\BIBentryALTinterwordspacing
K.~Sims, ``Artificial evolution for computer graphics,'' in \emph{Computer
  Graphics}, vol.~25, no.~4, ACM SIGGRAPH.\hskip 1em plus 0.5em minus
  0.4em\relax New York: ACM SIGGRAPH, July 1991, pp. 319--328. [Online].
  Available: \url{http://www.genarts.com/karl/papers/siggraph91.html}
\BIBentrySTDinterwordspacing

\bibitem{McCormack1992}
J.~McCormack, ``Interactive evolution of forms,'' in \emph{Cultural Diversity
  in the Global Village: Third International Symposium on Electronic Art},
  A.~Cavallaro, R.~Harley, L.~Wallace, and M.~Wark, Eds.\hskip 1em plus 0.5em
  minus 0.4em\relax Sydney, Australia: The Australian Network for Art and
  Technology, 1992, p. 122.

\bibitem{Todd1992}
S.~Todd and W.~Latham, \emph{Evolutionary Art and Computers}.\hskip 1em plus
  0.5em minus 0.4em\relax London: Academic Press, 1992.

\bibitem{Takagi2001}
H.~Takagi, ``Interactive evolutionary computation: Fusion of the capabilities
  of {EC} optimization and human evaluation,'' \emph{Proceedings of the IEEE},
  vol.~89, pp. 1275--1296, Sep 2001.

\bibitem{Secretan2011}
J.~Secretan, N.~Beato, D.~B. D'Ambrosio, A.~Rodriguez, A.~Campbell, J.~T.
  Folsom-Kovarik, and K.~O. Stanley, ``Picbreeder: A case study in
  collaborative evolutionary exploration of design space,'' \emph{Evolutionary
  Computation}, vol.~19, no.~3, pp. 373--403, 2011.

\bibitem{bownMcCormack2010}
O.~Bown and J.~McCormack, ``Taming nature: tapping the creative potential of
  ecosystem models in the arts,'' \emph{Digital Creativity}, vol.~21, no.~4,
  pp. 215--231, 2010.

\bibitem{Johnson2019}
\BIBentryALTinterwordspacing
C.~G. Johnson, J.~McCormack, I.~Santos, and J.~Romero, ``Understanding
  aesthetics and fitness measures in evolutionary art systems,''
  \emph{Complexity}, vol. 2019, no. Article ID 3495962, p. 14 pages, 2019.
  [Online]. Available: \url{https://doi.org/10.1155/2019/3495962}
\BIBentrySTDinterwordspacing

\bibitem{McCormack2005a}
J.~McCormack, ``Open problems in evolutionary music and art,'' in
  \emph{EvoWorkshops}, ser. Lecture Notes in Computer Science, F.~Rothlauf,
  J.~Branke, S.~Cagnoni, D.~W. Corne, R.~Drechsler, Y.~Jin, P.~Machado,
  E.~Marchiori, J.~Romero, G.~D. Smith, and G.~Squillero, Eds., vol.
  3449.\hskip 1em plus 0.5em minus 0.4em\relax Springer, 2005, pp. 428--436.

\bibitem{Leder2004}
H.~Leder, B.~Belke, A.~Oeberst, and D.~Augustin, ``A model of aesthetic
  appreciation and aesthetic judgments,'' \emph{British Journal of Psychology},
  vol.~95, pp. 489--508, 2004.

\bibitem{Leder2014}
H.~Leder and M.~Nadal, ``Ten years of a model of aesthetic appreciation and
  aesthetic judgments: The aesthetic episode -- developments and challenges in
  empirical aesthetics,'' \emph{British Journal of Psychology}, vol. 105, pp.
  443--464, 2014.

\bibitem{Baluja1994}
S.~Baluja, D.~Pomerleau, and T.~Jochem, ``Simulating user's preferences:
  Towards automated artificial evolution for computer generated images,''
  \emph{Connection Science}, vol.~6, pp. 325--354, 1994.

\bibitem{McCormackLomas2020}
J.~McCormack and A.~Lomas, ``Understanding aesthetic evaluation using deep
  learning,'' in \emph{Artificial Intelligence in Music, Sound, Art and Design.
  EvoMUSART 2020}, ser. LNCS, J.~Romero, A.~Ek{\'a}rt, T.~Martins, and
  J.~Correia, Eds.\hskip 1em plus 0.5em minus 0.4em\relax Springer, Cham, 2020,
  vol. 12103.

\bibitem{McCormackLomas2020b}
------, ``Deep learning of individual aesthetics,'' \emph{Neural Computing and
  Applications}, vol.~33, pp. 3--17, October 2021.

\bibitem{tutum2018functional}
C.~C. Tutum, S.~Chockchowwat, E.~Vouga, and R.~Miikkulainen, ``Functional
  generative design: an evolutionary approach to {3D}-printing,'' in
  \emph{Proceedings of the Genetic and Evolutionary Computation Conference},
  2018, pp. 1379--1386.

\bibitem{yu2017evolutionary}
E.~A. Yu, J.~Yeom, C.~C. Tutum, E.~Vouga, and R.~Miikkulainen, ``Evolutionary
  decomposition for {3D} printing,'' in \emph{Proceedings of the Genetic and
  Evolutionary Computation Conference}, 2017, pp. 1272--1279.

\bibitem{zhao2016connected}
H.~Zhao, F.~Gu, Q.-X. Huang, J.~Garcia, Y.~Chen, C.~Tu, B.~Benes, H.~Zhang,
  D.~Cohen-Or, and B.~Chen, ``Connected fermat spirals for layered
  fabrication,'' \emph{ACM Transactions on Graphics (TOG)}, vol.~35, no.~4, pp.
  1--10, 2016.

\bibitem{barlow1989differential}
P.~Barlow, P.~Brain, and J.~Adam, ``Differential growth and plant tropisms: a
  study assisted by computer simulation,'' in \emph{Differential Growth in
  Plants}.\hskip 1em plus 0.5em minus 0.4em\relax Elsevier, 1989, pp. 71--83.

\bibitem{ball1999self}
P.~Ball, \emph{The self-made tapestry: pattern formation in nature}.\hskip 1em
  plus 0.5em minus 0.4em\relax Oxford University Press Oxford, 1999.

\bibitem{hansen2001completely}
N.~Hansen and A.~Ostermeier, ``Completely derandomized self-adaptation in
  evolution strategies,'' \emph{Evolutionary computation}, vol.~9, no.~2, pp.
  159--195, 2001.

\bibitem{forsythe2011predicting}
A.~{Forsythe}, M.~{Nadal}, N.~{Sheehy}, C.~J. {Cela-Conde}, and M.~{Sawey},
  ``Predicting beauty: fractal dimension and visual complexity in art.''
  \emph{British Journal of Psychology}, vol. 102, no.~1, pp. 49--70, 2011.

\bibitem{Shamos1978computational}
\BIBentryALTinterwordspacing
M.~I. Shamos, ``Computational geometry.'' \emph{Ph. D. Thesis, Yale
  University}, 1978. [Online]. Available:
  \url{https://ci.nii.ac.jp/naid/10000034927/en/}
\BIBentrySTDinterwordspacing

\bibitem{McCormack2021_DLADataset}
\BIBentryALTinterwordspacing
J.~McCormack and C.~C. Gambardella, ``{DLA Form Generation dataset},'' 1 2021.
  [Online]. Available:
  \url{https://bridges.monash.edu/articles/dataset/DLA_Form_Generation_dataset/13663400}
\BIBentrySTDinterwordspacing

\bibitem{Lakhal2020}
\BIBentryALTinterwordspacing
S.~Lakhal, A.~Darmon, J.-P. Bouchaud, and M.~Benzaquen, ``Beauty and structural
  complexity,'' \emph{Phys. Rev. Research}, vol.~2, p. 022058, Jun 2020.
  [Online]. Available:
  \url{https://link.aps.org/doi/10.1103/PhysRevResearch.2.022058}
\BIBentrySTDinterwordspacing

\bibitem{McCormack2021Enigma}
\BIBentryALTinterwordspacing
J.~McCormack, C.~C. Gambardella, and A.~Lomas, ``The enigma of complexity,''
  \emph{arXiv preprint arXiv:2102.02332}, 2021. [Online]. Available:
  \url{https://arxiv.org/abs/2102.02332}
\BIBentrySTDinterwordspacing

\bibitem{Deb2002}
K.~Deb, A.~Pratap, S.~Agarwal, and T.~Meyarivan, ``A fast and elitist
  multiobjective genetic algorithm: {NSGA-II},'' \emph{IEEE Transactions on
  Evolutionary Computation}, vol.~6, no.~2, pp. 182--197, April 2002.

\end{thebibliography}
%

\begin{IEEEbiography}[{\includegraphics[width=1in,height=1.25in,clip,keepaspectratio]{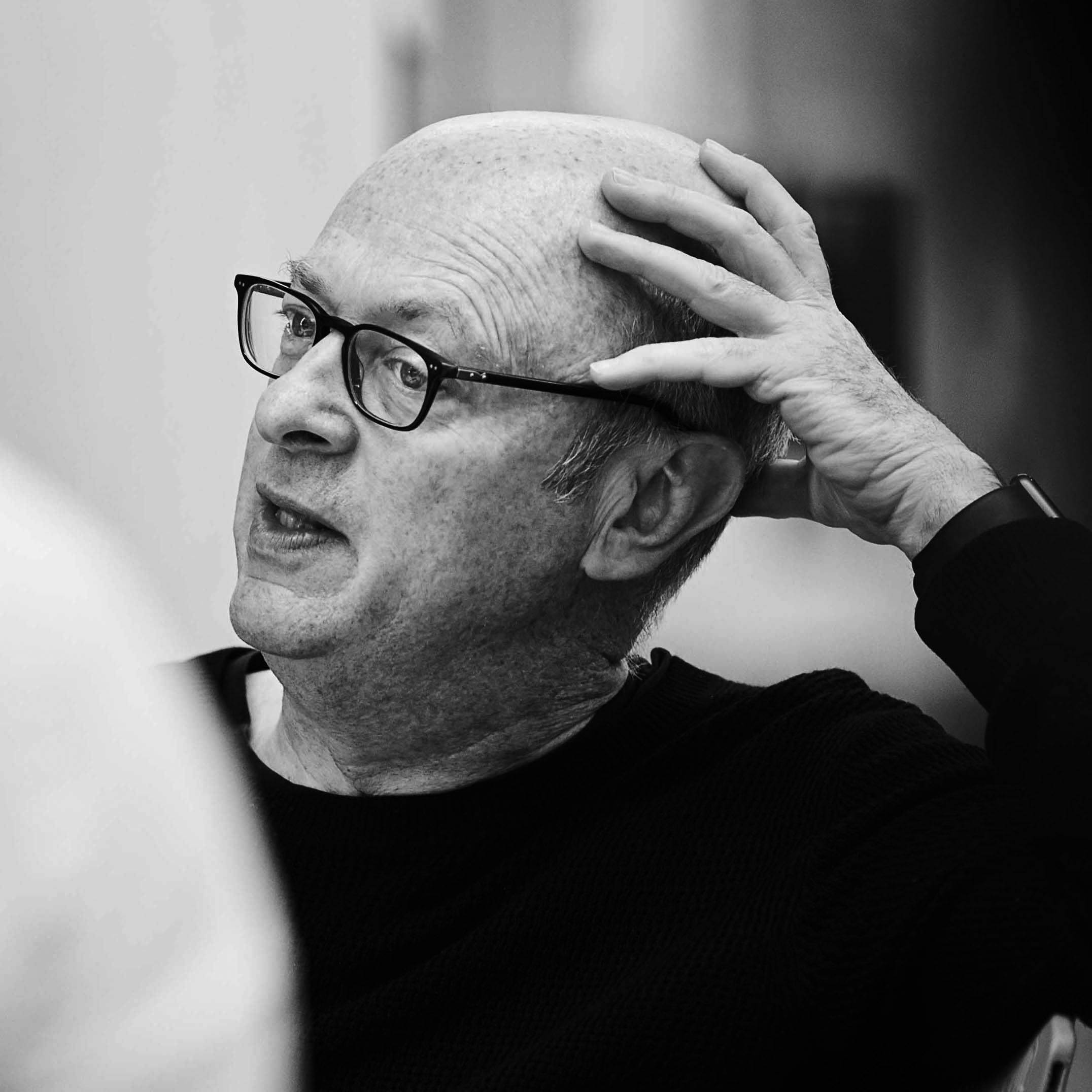}}]{Jon McCormack}
 is an Australian-based artist and researcher in computing. His research interests include generative art, design and music, evolutionary systems, computer creativity, developmental models and physical computing.
He is the founder and Director of SensiLab, a creative technologies research laboratory based at Monash University in Melbourne, Australia. He is also full Professor of Computer Science at Monash University and an Australian Research Council Future Fellow.
He holds an Honours degree in Applied Mathematics and Computer Science from Monash University, a Graduate Diploma of Art (Film and Television) from Swinburne University and a PhD in Computer Science from Monash University.
\end{IEEEbiography}
\begin{IEEEbiography}[{\includegraphics[width=1in,height=1.25in,clip,keepaspectratio]{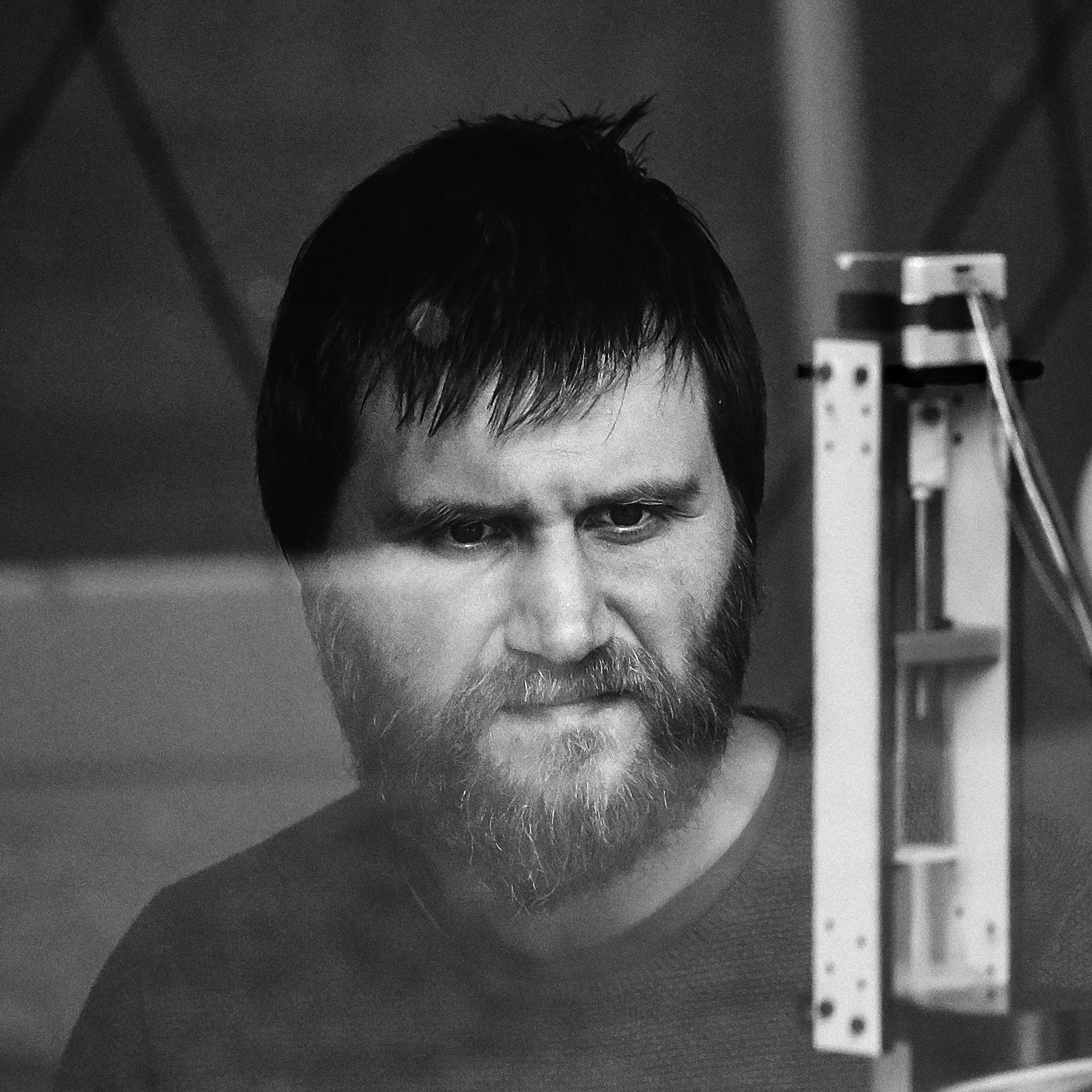}}]{Camilo Cruz Gambardella}
 is a research fellow at SensiLab, currently working on the development of generative and evolutionary methods for design and creativity using digital fabrication techniques.
Before coming to SensiLab, Camilo spent six years researching and teaching topics that sit in the intersection between design and computation at The University of Melbourne and The University of Chile. Prior to that, he also served as a visualisation artist and architectural designer in Toronto (Canada) and Santiago (Chile).
Camilo holds Bachelor and Professional degrees in Architecture from P. U. Catolica (Chile), a Master of Urban Design from the University of Toronto (Canada) and a PhD in Architecture and Computational Design from The University of Melbourne.
\end{IEEEbiography}


\vfill


\end{document}